\def\year{2020}\relax
\documentclass[letterpaper]{article} 
\usepackage{aaai20}  
\usepackage{times}  
\usepackage{helvet} 
\usepackage{courier}  
\usepackage[hyphens]{url}  
\usepackage{graphicx} 
\usepackage{natbib}
\usepackage{graphicx}  
\usepackage{amssymb}
\usepackage{boldline,multirow}
\usepackage{subfigure}
\usepackage{multicol}
\usepackage{appendix}
\title{TDAPNet: Prototype Network with Recurrent Top-Down Attention for Robust Object Classification under Partial Occlusion}

\begin{document}

\author{Mingqing Xiao$^{1\footnotemark[2]}$, Adam Kortylewski$^2$, Ruihai Wu$^{1\footnotemark[2]}$, Siyuan Qiao$^2$, Wei Shen$^2$, Alan Yuille$^2$ \\ $^1$Peking University, $^2$Johns Hopkins University \\ \{mingqing\_xiao, wuruihai\}@pku.edu.cn, \{akortyl1,siyuan.qiao,wshen10,alan.yuille\}@jhu.edu}

\maketitle

\renewcommand{\thefootnote}{\fnsymbol{footnote}}
\footnotetext[2]{Work done at Johns Hopkins University}


\begin{abstract}
Despite deep convolutional neural networks' great success in object classification, it suffers from severe generalization performance drop under occlusion due to the inconsistency between training and testing data. Because of the large variance of occluders, our goal is a model trained on occlusion-free data while generalizable to occlusion conditions. In this work, we integrate prototypes, partial matching and top-down attention regulation into deep neural networks to realize robust object classification under occlusion. We first introduce prototype learning as its regularization encourages compact data clusters, which enables better generalization ability under inconsistent conditions. Then, attention map at intermediate layer based on feature dictionary and activation scale is estimated for partial matching, which sifts irrelevant information out when comparing features with prototypes. Further, inspired by neuroscience research that reveals the important role of feedback connection for object recognition under occlusion, a top-down feedback attention regulation is introduced into convolution layers, purposefully reducing the contamination by occlusion during feature extraction stage. Our experiment results on partially occluded MNIST and vehicles from the PASCAL3D+ dataset demonstrate that the proposed network significantly improves the robustness of current deep neural networks under occlusion. Our code will be released.
\end{abstract}

\begin{figure} [htbp]
\centering
\includegraphics[scale=0.26]{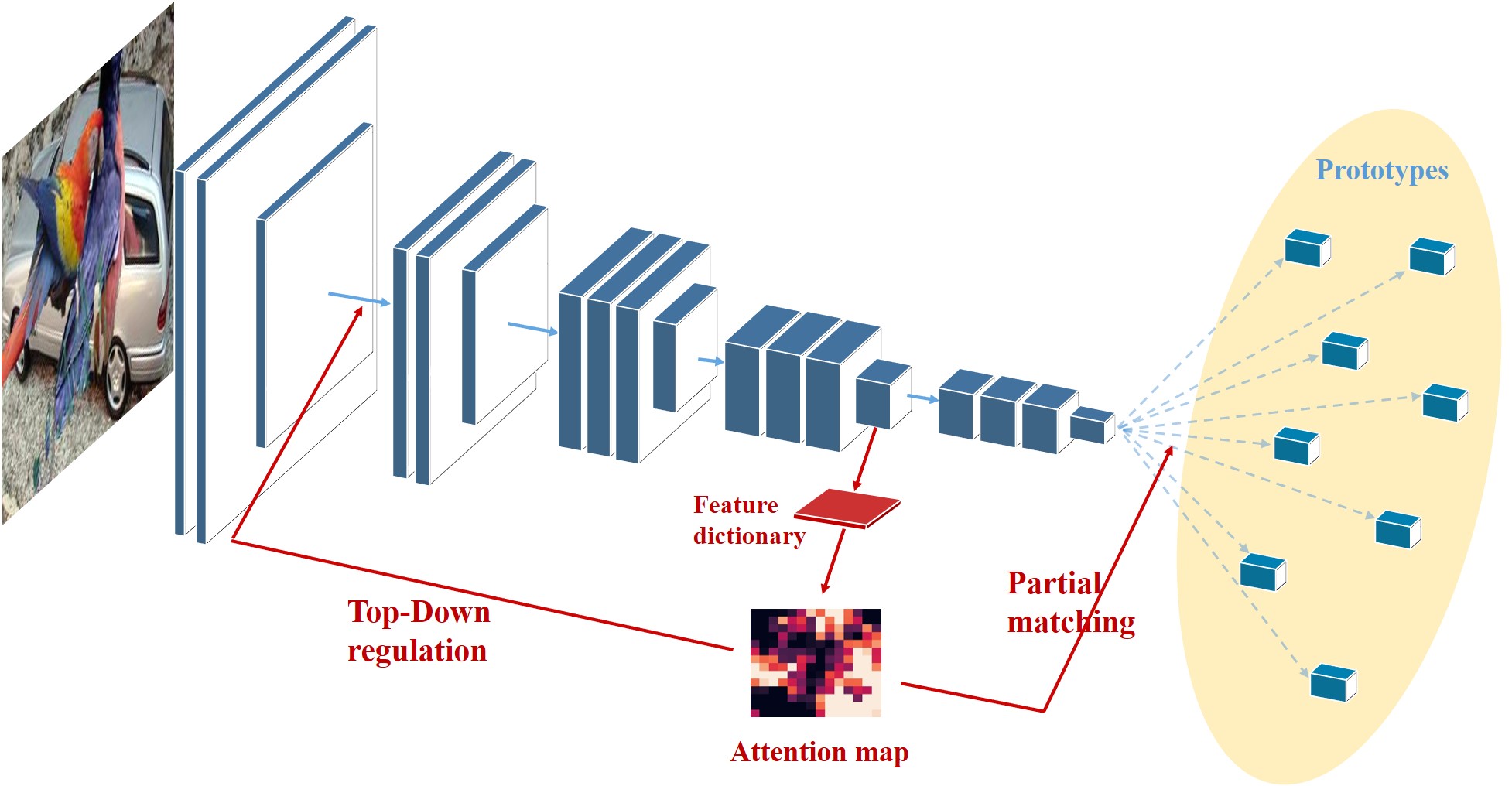}
\caption{Overall architecture of TDAPNet. We use the convoluation layers of VGG-16 as our feature extractor and conduct prototype matching on the features. We estimate an attention map from the pool-4 layer to focus on informative parts. The attention map is first used for top-down regulation, reducing the contamination of occlusion during feature extraction, and then for partial matching, sifting irrelevant information out when comparing features and prototypes.}
\label{architecture}
\end{figure}

\section{Introduction}
In recent years, deep convolutional neural networks (DCNNs) have achieved great success in computer vision tasks, like image classification \citep{sutskever2012imagenet,simonyan2014very,he2016deep} and object detection \citep{ren2015faster,redmon2016you}. However, widely used deep learning models are not robust under occlusion conditions, even with artificial masks \citep{BMVC2016_137,wang2017visual,zhang2018deepvoting,kortylewski2019compositional}. Human, on the other hand, are still able to recognize recognize objects under extreme occlusions by masks or other objects \citep{zhu2019robustness}. Therefore, a reliable computer vision model must be robust to recognize objects under occlusion of masks and objects, and be able to prevent such kind of attack.

A common hypothesis is that the distribution inconsistency between clean training data and occluded testing data mainly leads to the failure of traditional DCNNs \citep{kortylewski2019compositional}. DCNNs first extract features through blocks of convolution layers and then do the classification in the feature space. The learned classifier, however, may not generalize well due to the contamination caused by occlusion or masks. Because occlusion mask patterns are highly variable in terms of apprearance and shape, including all possible patterns in the training data is impossible. And only biased occlusion patterns in training data cannot improve much generalization performance in other conditions \citep{kortylewski2019compositional}. Therefore, our work focuses on training a model on occlusion-free data while generalizable to occlusion conditions without assumptions on occlusion.

There are two main challenges. The first one is the over-fitting on the training dataset, which reduces the generalization ability under occlusion conditions. Another one is that contamination in the surroundings around occlusion disturbs the feature extraction. We introduce prototype learning and partial matching to deal with the first problem. And a top-down attention regulation is designed to tackle the second problem.

In cognitive science, prototype-matching is a popular theory that describes how human recognize objects --- our brains compare the stimulus to prototypes and an object is recognized if a similar prototype match is found. From mathematical perspective, prototypes can be viewed as cluster centers of points from the same class in an embedding space, and distance performs as the matching function. Prototype learning after feature extraction is able to deal with over-fitting \citep{snell2017prototypical}, as it imposes regularization with nearest neighbor inductive bias to encourage compact data clusters. Furthermore, different prototypes in one class are able to account for large changes in spatial patterns, such as different viewpoints for 3D objects \citep{kortylewski2019compositional}. Prototypes have been introduced and integrated into deep network structure in few-shot learning task \citep{snell2017prototypical} and for rejection and class-incremental learning \citep{yang2018robust}. However, their distances are simply euclidean distances, which cannot be directly used in occlusion conditions due to the contamination of occlusion. Experiments demonstrate that pure prototype learning improves DCNNs slightly.

To tackle the problem of prototype matching under occlusion conditions, we introduce partial matching based on the attention that focuses on informative parts, as the attention map shown in Figure \ref{architecture}. Since occlusion or masks are hard to predict, only unoccluded informative parts should be focused on. \citet{wang2015unsupervised} first discovered that semantic part representations for objects can be found from the internal states of trained DCNNs, based on which \citet{wang2017visual} and \citet{zhang2018deepvoting} developed semantic part detection methods. Besides, larger activation scales of internal states are also correlated with objects \citep{zhang2018interpretable}. We estimate attention by finding possible object parts from internal DCNN states. Experiemnts show that partial matching according to the estimated attention is able to sift occlusion out effectively.

In addition, we propose a top-down feedback regulation with the estimated part attention, as the feedback connection shown in Figure \ref{architecture}, because occluders could also contaminate surrounding features during the feature extraction stage. The feedback attention could help the bottom layers to filter occlusion-caused anomalous activitions with high-level information, so that areas around the occluders suffer less from contamination. Experiments in section 4 demonstrate the effective contamination reduction. Our top-down regulation is related with some neuroscience conjectures. There're several neuroscience evidence show that recurrent and feedback connections play an important role in object recognition when stimuli are partially occluded \citep{gilbert2013top,o2013recurrent,spoerer2017recurrent,rajaei2019beyond}. The main conjectures include that the recurrence fills missing data and that it sharpens certain representations by attention refinement \citep{nayebi2018task}. Here we assume that top-down connection could be a neural regulation of attention to filter occlusion-caused anomalous activations. Our top-down feedback attention regulation is purposefully to reduce the contamination of occlusion during feature extraction and is composed of interpretable informative-part attention.

In summary, this paper makes the following contributions:

\begin{itemize}
\item We introduce prototype learning and partial matching with informative-part attention for robust object classification under occlusion. The prototypes and attention are integrated into the neural network. Therefore, the whole structure is end-to-end and about the same amount of parameters as neural networks.
\item We further propose a top-down feedback regulation based on the part attention in convolution layers. It imitates the neurological regulation from higher cortex to lower cortex, and is explicitly meaningful. Experiments show that the feedback effectively reduce the contamination of occlusion during feature extraction.
\item Experiments on PASCAL3D+ and MNIST demonstrate that the proposed model significantly improves the robustness of DCNNs with increase of 11\% on PASCAL3D+ and 17.2\% on MNIST for average object classification accuracy under different occlusion conditions.
\end{itemize}

\section{Related Work}
\subsection{Object classification under partial occlusion}

\citet{BMVC2016_137} have shown that DCNNs are not robust to partial occlusion when inputs are masked out by patches. \citet{devries2017improved} and \citet{yun2019cutmix} proposed regularization methods by masking out patches from the images during training, but these approaches have not shown improvements on the robustness to partial occlusion at test time. \citet{kortylewski2019compositional} proposed dictionary-based Compositional Model. Their model is composed of two parts: the first is a traditional DCNN and the second is a compositional model based on the features extracted by DCNN. At runtime, the input is first classified by the DCNN, and will turn to compositional model only when the prediction uncertainty exceeds a threshold, because compositional models are less discriminative than DCNNs. Their model is not end-to-end, does not consider contamination of occlusion during feature extraction and requires a model of occluders. Differently, our proposed model follows the deep network architecture, reduces influence of occlusions both during and after feature extraction, and is generalizable without information for occluders.

\subsection{Prototype learning in deep networks}

Prototype learning is a classical method in pattern recognition. After the rise of deep neural networks, \citet{yang2018robust} replace the traditional hand-designed features with features extracted by convolutional neural networks in prototype learning and integrate it into deep networks for both high accuracy and robust pattern classification. Prototypes are also introduced in few-shot and zero-shot learning as part of metric learning \citep{vinyals2016matching, snell2017prototypical}. Nevertheless, all these works use basic measures like euclidean or cosine distance in prototype matching, which is not suitable for occlusion conditions. We introduce the informative-part attention to extend prototype matching to occlusion conditions.

\subsection{Object part representation inside DCNNs}

\citet{wang2015unsupervised} found that by clustering feature vectors at different positions from the intermediate layer of a pre-trained deep neural network, e.g. pool-4 layer in VGG, the patterns of some cluster centers, which are called ``visual concepts", are able to reflect specific object parts. \citet{wang2017visual} and \citet{zhang2018deepvoting} use it for semantic part detection, and \citet{kortylewski2019compositional} use it to obtain part components in their compositional model. Related works also include \citet{liao2016learning}, which added a regularizer to encourage the feature representations of DCNNs to cluster during learning, trying to obtain part representations. From another perspective, \citet{zhang2018interpretable} tried to encourage each filter to be a part detector by restricting the activations of each filter to be independent, and they estimated the part position by the activation scale. These works demonstrate that object part representation could be obtained inside DCNNs, and activation scale contains information. Based on these ideas, we obtain attention map for partial prototype matching under occlusion by finding parts that are informative for classification, e.g. object parts, with their representations and activation scales.

\subsection{Feedback connections in deep networks}

Despite top-down feedback connection is an ubiquitous structure in biological vision systems, it is not used in typical feed-forward DCNNs. \citet{nayebi2018task} has summarized the function conjectures of recurrence in the visual systems and explored possible recurrence structures in CNNs to improve classification performance through architecture search. As for classification task under occlusion, \citet{spoerer2017recurrent} explored top-down and lateral connections for digit recognition under occlusion, but their function is not explicit. As for top-down attention information, \citet{fu2017look} learned to focus on smaller areas in the image and \citet{li2018learning} designed a feedback layer and an emphasis layer. But all of their attention layers are composed of fully connected layers, which is not interpretable. Some DCNN architectures also borrow the top-down feedback idea, like CliqueNet \citep{yang2018convolutional}. Different from these works, our top-down feedback attention regulation is composed of explainable part attention and is purposefully for reduction in contamination of occlusion.

\section{Method}

Our model is composed of three main parts. The first is prototype learning after feature extraction. Following it is partial matching based on estimated informative-part attention in order to extend prototype matching under occlusion. Finally, top-down regulation according to the part attention is introduced to reduce the contamination of occlusion.

\subsection{Prototype learning}

We conduct prototype learning after feature extraction by DCNNs. Let $x \in \mathbb{R} ^ {H_0 \times W_0 \times 3}$ denote the input image, our feature extractor is $f_\theta: \mathbb{R}^{H_0 \times W_0 \times 3} \to \mathbb{R}^{H \times W \times C}$, which is composed of convolution layer blocks in typical DCNNs. In contrast to related works \citep{snell2017prototypical, yang2018robust}, our feature is a tensor $f_\theta(x) \in \mathbb{R} ^ {H \times W \times C}$ rather than a vector, so that it can maintain spatial information of objects for partial matching in next section. Suppose there are $N$ classes for classification, we set $M$ prototypes for each class to account for differences in spatial activation patterns. Therefore prototypes are a set of tensors $p_{i,j} \in \mathbb{R} ^ {H \times W \times C}$, where $i \in \{1,2,...,N\}$ denotes the class the prototype belongs to, and $j \in \{1,2,...,M\}$ represents the index of the prototype in its class.

For feedforward prediction, the image is classified to the class of its nearest prototype according to a distance function $d: \mathbb{R}^{H \times W \times C} \times \mathbb{R}^{H \times W \times C} \to [0,+\infty)$: 

\begin{equation}
    Pred(x) = \mathop{\arg\min}_{i} \{\mathop{\min}_{j} d(f_\theta(x),p_{i,j})\}.
    \label{pred}
\end{equation}

The distance function $d$ may simply be euclidean distance, but our experiments will show that it improves networks slightly due to the contamination of occlusion. A distance for partial matching will be introduced next section.

For backward update of parameters, we use cross entropy loss based on the distances. To be specific, distances between the feature $f_\theta(x)$ and prototypes $p_{i,j}$ produce a probability distribution over classes:

\begin{equation}
    Pr(y=k|x) = \frac{\exp(-\gamma d_k)}{\sum_{i=1}^{N}\exp(-\gamma d_i)},
    \label{probability}
\end{equation}

\noindent where $d_k=\mathop{\min}_{j} d(f_\theta(x),p_{k,j})$, and $\gamma$ is a parameter that control the hardness of probability assignment. We set $\gamma$ to be learned by network automatically. Then based on the probability, cross entropy loss is defined:

\begin{equation}
    L_{ce}((x,k);\theta,\{p_{i,j}\}) = -\log Pr(y=k|x).
\end{equation}

Further, a prototype loss is added to act as the regularization of prototype learning:

\begin{equation}
    L_p((x,k);\theta,\{p_{i,j}\}) = \mathop{\min}_{i,j} d(f_\theta(x), p_{i,j}).
    \label{prloss}
\end{equation}

Different from \citet{yang2018robust}, we only consider the nearest prototype when computing distances and probabilities, because our $M$ prototypes in the same class are designed to represent different states of objects, such as different viewpoints, which may vary a lot in spatial distribution.

We initialize the prototypes by clustering the features of a sub dataset using k-means algorithm \citep{lloyd1982least}. It prevents the degeneration of multiple prototypes to a single prototype.


\subsection{Partial matching under occlusion}

The core problem for extending prototype learning directly to occlusion conditions is the matching function. Since occlusion will contaminate the object feature representation, simple distance between the feature and prototypes won't be valid enough to do classification. Experiments will show that pure prototype matching improves deep neural networks slightly. An attention for valid parts in features is required.

We estimate an attention map based on feature dictionary and activation scale to focus on valid unoccluded parts in features, which enables partial matching. We learn a feature dictionary in the intermediate by clustering feature vectors over the whole dataset on the feature map, which can represent specific activation patterns of parts in the images. Specifically, feature dictionary is obtained by clustering all normalized vector $v_{k,i,j} \in \mathbb{R} ^ {1 \times 1 \times C^l}$ at position $(i, j)$ of the feature map $f_\theta^l(x_k) \in \mathbb{R} ^ {H^l \times W^l \times C^l}$ at the intermediate layer $l$ over the dataset \{$x_k$\}. Related works \citep{wang2015unsupervised,kortylewski2019compositional} show that cluster centers are mostly activated by similar parts in the images, most of which are object parts. For more detailed visualization, refer to related works \citep{wang2015unsupervised, wang2017visual, kortylewski2019compositional}. Based on the feature dictionary \{$d_k$\}, we compare the similarity between the vectors $f_\theta^l(x)_{i,j}$ of the feature at layer $l$ and each component $d_k$: $S(f_\theta^l(x)_{i,j}, d_k)=\frac{f_\theta^l(x)_{i,j}}{\left\|f_\theta^l(x)_{i,j}\right\|_2} \cdot d_k$. The higher the maximum similarity over \{$d_k$\} is, the more likely is this area a specific object part. Therefore, we can sift occlusion out due to its low similarity.

\begin{figure} [tp]
    \centering
    \subfigure[]{
    \includegraphics[scale=0.29]{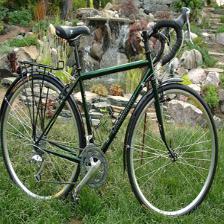}
    \label{scale_1}
    }
    \subfigure[]{
    \includegraphics[scale=0.134]{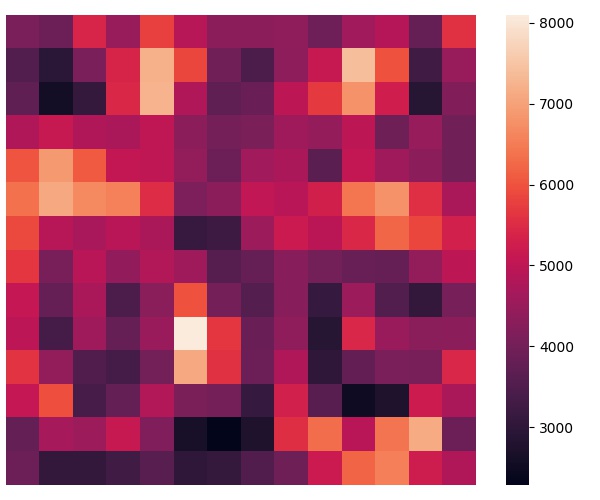}
    \label{sacle_1_scale}
    }
    \subfigure[]{
    \includegraphics[scale=0.29]{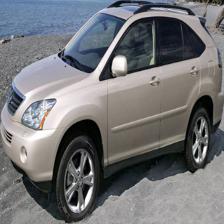}
    \label{scale_2}
    }
    \subfigure[]{
    \includegraphics[scale=0.134]{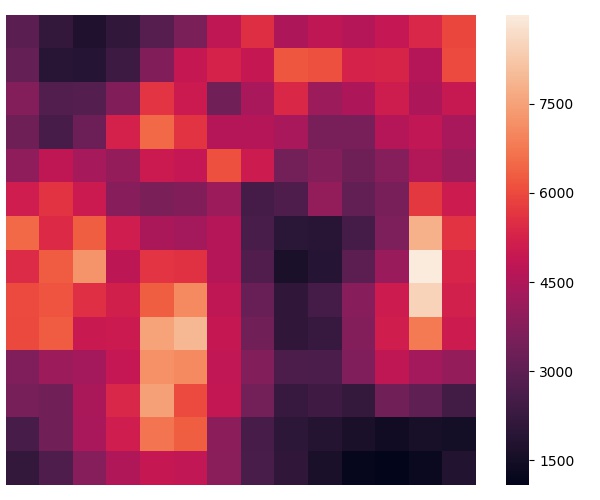}
    \label{sacle_2_scale}
    }
    \caption{Visualizaion of activation scale in the pool-4 layer. It shows that the activation scale at parts informative for classification, e.g. object parts, is larger than other areas.}
    \label{activation_scale}
\end{figure}

However, there are also a few background activation patterns irrelevant to classification in the dictionary. We use the relative scale of activations to filter them out. As shown in the Figure \ref{activation_scale}, the scales of activations in a trained network for most irrelevant background are much lower than objects. It is probably because deep networks could learn to focus on image parts that contribute to discrimination most. Considering activation scales is helpful to filter irrelevant background and maintain most informative signals.

Combining the similarity with the feature dictionary and the activation scale enables us to estimate an attention map that focuses on unoccluded informative parts. The formulation for attention at postion $(i, j)$ in layer $l$ is:

\begin{equation}
    a_{i,j}^l = ReLU(\mathop{\max}_k f_\theta(x)^l_{i,j} \cdot d_k).
\end{equation}

Since the scale of the activation after the ReLU function could be large, we normalize $a_{i,j}^l$. We use the following linear function with clipping since it preserves proper relative relationship among activation scales:

\begin{equation}
\label{threshold}
    A_{i,j}^l = \frac{\min(\max(a_{i,j}^l,a_l), a_u)}{a_u},
\end{equation}

\noindent where $a_l$ and $a_u$ are lower and upper thresholds that can be dynamically determined according to \{$a_{i,j}^l$\}.

Subsequently, the attention map is down-sampled to the same spatial scale of $f_\theta(x)_{i,j}$ for partial matching. We denote it as $\{A_{i,j}\}$. Based on \{$A_{i,j}$\}, partial matching between the feature and prototypes is enabled. Let $f_\theta(x) \odot A$ denote the application of attention by scaling vectors $f_\theta(x)_{i,j}$ with $A_{i,j}$. A distance for partial matching under occlusion, which is used for Eq.(\ref{pred}), (\ref{probability}) and (\ref{prloss}), is defined as:

\begin{equation}
    d(f_\theta(x),p_{i,j}) = \frac{1}{2} \left\|f_\theta(x) \odot A - p_{i,j} \odot A \right\|_2^2
\end{equation}

In this way, we only compare unoccluded object parts based on the estimated attention map. Due to the high-dimension of $f_\theta(x)$ and $p_{i,j}$, we normalize them on a unit sphere at first and compute the euclidean distance after applying attention, in order to obtain a valid distance.

We learn the feature dictionary \{$d_k$\} through clustering. So similar to prototype learning, we initialize it with clustering result on the pre-trained neural network, and add the clustering loss in the whole loss function during training:

\begin{equation}
    L_d = \sum_{i,j} \mathop{\min}_k \frac{1}{2} \left\|\frac{f_\theta(x)_{i,j}}{\left\|f_\theta(x)_{i,j}\right\|_2} - d_k\right\|_2^2
\end{equation}

Note that we simply add a normalization layer in the network to normalize $d_k$ and ignore the notation in the formula.

\subsection{Top-Down feedback regulation based on part attention}

Our proposed partial matching only sifts out irrelevant feature vectors when comparing features and prototypes. However, occluders may also contaminate its nearby feature vectors. We propose to filter the occlusion-caused anomalous activitions in the lower layers to reduce such contamination and thus obtain cleaner features around the occluder.

Based on the estimated attention map at a higher layer, a top-down feedback connection is introduced to reduce the contamination of occlusion in lower layers. Formally, denote \{$A_{i,j}^b$\} as the up-sampling attention result of \{$A_{i,j}^l$\} to the same spatial size as the bottom layer b, such as pool-1 layer, and $f_\theta^b$ as the function from input to layer b. A new activation pattern at layer b can be obtained by applying the attention to the old activation:

\begin{equation}
    f_\theta^b(x)_{new} = f_\theta^b(x) \odot A^b
\end{equation}

\begin{table*} [ht]
	\small
	\centering
	\tabcolsep=0.11cm
	\begin{tabular}{lV{2.5}cV{2.5}c|c|c|cV{2.5}c|c|c|cV{2.5}c|c|c|cV{2.5}c}
		\multicolumn{15}{c}{\textbf{PASCAL3D+ Classification under Occlusion}} \\
		\hline
		Occ. Area 					         & \textbf{0\%} & \multicolumn{4}{cV{2.5}}{\textbf{Level-1: 20-40\%}} & \multicolumn{4}{cV{2.5}}{\textbf{Level-2: 40-60\%}} & \multicolumn{4}{cV{2.5}}{\textbf{Level-3: 60-80\%}} & Mean \\
		\hline
		Occ. Type 	& - & w & n & t & o & w & n & t& o & w & n & t & o &-\\    		
		\hline  
		\hline
		VGG 		& \textbf{99.4} &97.5&97.5&97.3&92.1&91.7&90.6&90.2&73.0&65.0&60.7&56.4&52.2&81.8\\		
		\hline
		CompDictModel \citep{kortylewski2019compositional}		&98.3&96.8&95.9&96.2&94.4&91.2&91.8&91.3&91.4&71.6&80.7&77.3&\textbf{87.2}&89.5\\
		\hline		
		\hline
		PrototypeNet without partial matching
		&99.2&97.1&97.6&97.2&95.3&91.2&93.0&91.3&81.3&61.9&60.9&57.9&61.5&83.5\\
		\hline					
		PrototypeNet with partial matching
		&99.3&98.4&\textbf{98.9}&98.5&97.3&\textbf{96.4}&97.1&96.2&89.2&\textbf{84.0}&87.4&79.7&74.5&92.1\\
		\hline
		TDAPNet with 1 recurrence
		&99.3&98.4&\textbf{98.9}&\textbf{98.7}&97.2&96.1&97.4&96.4&90.2&81.1&87.6&81.2&76.8&92.3\\
		\hline
		TDAPNet with 2 recurrence
		&99.2&\textbf{98.5}&98.8&98.5&97.3&96.2&97.4&\textbf{96.6}&90.2&81.5&87.7&81.9&77.1&92.4\\
		\hline
		TDAPNet with 3 recurrence
		&99.3&98.4&\textbf{98.9}&98.5&\textbf{97.4}&96.1&\textbf{97.5}&\textbf{96.6}&\textbf{91.6}&82.1&\textbf{88.1}&\textbf{82.7}&79.8&\textbf{92.8}\\
		\hline
		TDAPNet with 4 recurrence
		&99.3&98.4&\textbf{98.9}&98.4&97.2&96.0&\textbf{97.5}&96.5&91.4&81.5&87.7&82.4&79.3&92.7\\
		\hline		
		\hline		
		Human \citep{kortylewski2019compositional} & 100.0& \multicolumn{4}{cV{2.5}}{100.0}& \multicolumn{4}{cV{2.5}}{100.0}  & \multicolumn{4}{cV{2.5}}{98.3}& 99.5
		\vspace{.2cm}
	\end{tabular}
	\begin{tabular}{lV{2.5}cV{2.5}c|c|cV{2.5}c|c|cV{2.5}c|c|cV{2.5}c}
		\multicolumn{12}{c}{\textbf{MNIST Classification under Occlusion}} \\
		\hline
		Occ. Area 					         & \textbf{0\%} & \multicolumn{3}{cV{2.5}}{\textbf{Level-1: 20-40\%}} & \multicolumn{3}{cV{2.5}}{\textbf{Level-2: 40-60\%}} & \multicolumn{3}{cV{2.5}}{\textbf{Level-3: 60-80\%}} & Mean \\
		\hline
		Occ. Type 	& - & w & n & t & w & n & t & w & n & t &-\\    		
		\hline  
		\hline
		VGG 		& 99.4 &76.8&63.1&71.4&51.1&41.9&43.2&24.9&25.7&23.5&52.1\\		
		\hline
		CompDictModel \citep{kortylewski2019compositional}		&99.1&85.2&82.3&83.4&72.4&\textbf{71.0}&\textbf{72.8}&45.3&\textbf{41.2}&43.0&\textbf{69.4}\\
		\hline		
		\hline
		PrototypeNet without partial matching
		&99.3&81.0&71.8&77.4&53.4&44.4&50.4&27.4&28.3&29.9&56.3\\
		\hline					
		PrototypeNet with partial matching
		&99.4&86.3&78.8&82.9&67.3&56.1&59.7&43.6&36.8&37.6&64.9\\
		\hline
		TDAPNet with 1 recurrence
		&99.4&87.6&81.4&85.3&69.3&57.9&64.0&46.1&36.8&42.1&67.0\\
		\hline
		TDAPNet with 2 recurrence
		&99.4&88.2&82.2&85.5&70.6&59.8&64.9&47.0&38.8&42.8&67.9\\
		\hline
		TDAPNet with 3 recurrence
		&99.4&88.7&82.9&85.7&71.4&60.2&65.2&47.8&38.7&42.8&68.3\\
		\hline
		TDAPNet with 4 recurrence
		&\textbf{99.5}&\textbf{89.3}&\textbf{84.2}&\textbf{86.3}&\textbf{72.7}&61.6&66.3&\textbf{49.3}&40.0&\textbf{44.0}&69.3\\
		\hline		
		\hline		
		Human \citep{kortylewski2019compositional} & 100.0& \multicolumn{3}{cV{2.5}}{92.7}& \multicolumn{3}{cV{2.5}}{91.3}  & \multicolumn{3}{cV{2.5}}{64}& 84.4
		\vspace{.2cm}
	\end{tabular}
	\caption{Classification results for PASCAL3D+ and MNIST with different levels of occlusion (0\%, 20-40\%, 40-60\%, 60-80\% of the object are occluded) and different types of occlusion (w = white boxes, n = noise boxes, t = textured boxes, o = natural objects). PrototypeNet denotes only replacing fully-connected layers in VGG by prototype learning, without top-down attention. Without partial matching denotes simply use euclidean distance for prototype matching. All prototype numbers in one class are set to 4, to be the same as CompDictModel \citep{kortylewski2019compositional}. It shows that our model significantly outperforms VGG in every occlusion conditions, and performs especially well under low occlusion conditions compared with CompDictModel.}
	\label{classification}
\end{table*}

The new activation is again feed-forwarded, like a recurrent procedure. The recurrence can be carried out for multiple times, gradually refining features to reduce the contamination of occlusion. The upper threshold in eq.(\ref{threshold}) prevents degeneration of the attention to only one point, and the lower threshold in eq.(\ref{threshold}) prevents mistaken filtration due to the possible contamination of occlusion from the bottom layer to top layers.


In summary, the overall architecture with our three components is shown in Figure \ref{architecture}. Our overall loss function for training is:

\begin{equation}
    L = L_{ce}((x,k);\theta, \{p_{i,j}\}) + \lambda_1L_p((x,k);\theta, \{p_{i,j}\}) + \lambda_2L_d
    \label{loss}
\end{equation}




\section{Experiments}

\subsection{Dataset and settings}

\begin{figure} [ht]
    \centering
    \subfigure[]{
    \includegraphics[scale=0.4]{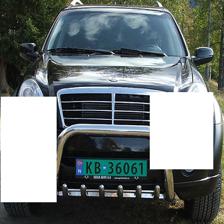}
    \label{p1w}
    }
    \subfigure[]{
    \includegraphics[scale=0.4]{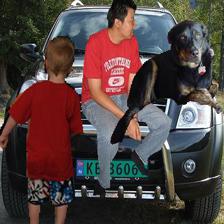}
    \label{p5o}
    }
    \subfigure[]{
    \includegraphics[scale=0.4]{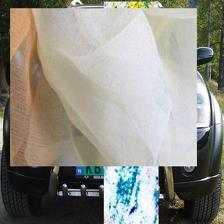}
    \label{p9t}
    }
    \quad
    \subfigure[]{
    \includegraphics[scale=0.3]{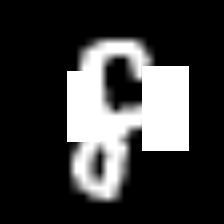}
    \label{m1w}
    }
    \subfigure[]{
    \includegraphics[scale=0.3]{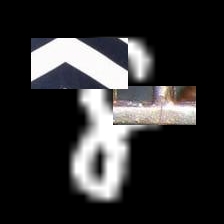}
    \label{m5t}
    }
    \subfigure[]{
    \includegraphics[scale=0.3]{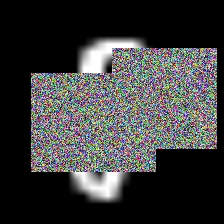}
    \label{m9n}
    }
    \caption{Examples of synthetic partial occlusion mask. (a), (b), (c) correspond to level 1-3 on PASCAL3D+; (d), (e), (f) correspond to level 1-3 on MNIST. Different types of occlusion appearances are: white boxes (a\&d), random noise (f), textures (c\&e), and natural objects (b).}
    \label{example}
\end{figure}

We evaluate our model for object classification on partially occluded MNIST digits \citep{lecun1998mnist} and vehicles from the PASCAL3D+ dataset \citep{xiang2014beyond}. Following \citet{kortylewski2019compositional}, we simulate partial occlusion by masking out patches in the images and filling them with white boxes, random noise or textures. In addition to the artificial box occluders, we also use the images provided in the VehicleSemanticPart dataset \citep{wang2017visual} for the PASCAL3D+ vehicles, where occlusion was simulated by superimposing segmented objects over the target object. The objects for partial occlusion are different from the objects for classification. Different occlusion levels are also defined corresponding to the percentage of occlusion over objects based on the object segmentation masks provided in the PASCAL3D+ and threshold segmentation of the MNIST digits. Examples refer to Figure \ref{example}. We use the standard splits for the train and test data. For the PASCAL3D+ dataset, we follow the setup in \citet{wang2015unsupervised} and \citet{kortylewski2019compositional}, that is the task is to discriminate between 12 objects during training, while at test time the six vehicle categories are tested.

We utilize convolution layers in a VGG-16 that was pre-trained on the ImageNet dataset as the feature extraction part of our model. Prototype learning is conducted on the pool-5 layer. The attention is estimated from the pool-4 layer, and the top-down regulation is imposed on pool-1 layer. The prototype number is specified in the results. We set feature dictionary components to be 512 for both datasets and use von Mises-Fisher clustering result \citep{kortylewski2019compositional} as the initialization. The thresholds in Eq.(\ref{threshold}) are dynamically determined: the upper threshold is the top 20\% value among \{$a_{i,j}^4$\}, and the lower threshold is the top 80\% value among \{$a_{i,j}^4$\}. The parameter $\gamma$ in Eq.(\ref{probability}) is initialized as 20, and the hyper-parameter $\lambda_1$ and $\lambda_2$ in Eq.(\ref{loss}) are simply set to 1. For PASCAL3D+, we first train prototypes individually for 10 epochs and then the whole network for 50 epochs; for MNIST, we first train prototypes individually for 10 epochs and then the whole network for 10 epochs. We use SGD with momemtum to train the model. Learning rate is 1e-3 for prototype training and 1e-4 for the whole network training. We use $l_2$ regularization for convolution layers. Batch normalization, data augmentation, or other regularization methods are not used. We compare our model with VGG-16 finetuned on the dataset, CompDictModel \citep{kortylewski2019compositional}, and human baseline. For VGG-16, the last layer is first finetuned for 5 epochs and the whole network is finetuned for 10 epochs until convergence. The regularization and optimization method for it is the same as our model. The result of CompDictModel and human baseline is directly from \citet{kortylewski2019compositional}. Both our model and VGG-16 are trained on non-occluded data and tested on occlusion conditions at different levels.

\subsection{Classification results}

Results for classification at different occlusion levels are shown in Table \ref{classification}. They show that DCNNs do not generalize well under partial occlusion, with a significant drop in accuracy. Our model significantly outperforms VGG in every occlusion conditions and remains the same accuracy when there's no occlusion.

\noindent\textbf{Pure prototype learning improves DCNNs slightly.} As shown in the results, direct prototype learning with simple distance function has little improvement. On PASCAL3D+, it mainly outperforms VGG under the condition when occlusions are objects. On MNIST, however, it outperforms VGG in most conditions, but the improvements are low compared with follow-up results.

\noindent\textbf{Partial matching plays a crucial role.} As illustrated by the results, partial matching significantly improves the performance. For the mean accuracy over all conditions, it improves 10.3 percent on PASCAL3D+ and 14.5 percent on MNIST compared with VGG. In the low occlusion level on PASCAL3D+, partial matching achieves the best results even without top-down attention.

\noindent\textbf{Top-Down attention works well for severe occlusions.} Top-down recurrence effectively improves the performance in relatively hard tasks that even human performance drops. As recurrence times goes up, the features are more pure and therefore performance increases. A more detailed analysis will be performed in the following section. With top-down attention, the finial mean accuracy outperforms VGG 11 percent on PASCAL3D+ and 17.2 percent on MNIST. This clearly reflects its robustness under partial occlusion.

\begin{table} [htbp]
	\small
	\centering
	\tabcolsep=0.11cm
	\begin{tabular}{lV{2.5}c|c|c|cV{2.5}c}
		\multicolumn{6}{c}{\textbf{PASCAL3D+ Classification under Occlusion}} \\
		\hline
		Occ. Area 	& \multicolumn{4}{cV{2.5}}{\textbf{Level-3: 60-80\%}} & Mean \\
		\hline
		Occ. Type 	& w & n & t & o &-\\    		
		\hline  
		\hline
		1 prototype, 1 recurrence
		&79.1&84.8&77.9&75.1&90.9\\
		\hline
		1 prototype, 2 recurrence
		&78.9&85.2&78.2&75.5&91.0\\
		\hline
		1 prototype, 3 recurrence
		&79.7&85.2&79&76.9&91.2\\
		\hline
		\hline
		4 prototype, 1 recurrence
		&81.1&87.6&81.2&76.8&92.3\\
		\hline
		4 prototype, 2 recurrence
		&81.5&87.7&81.9&77.1&92.4\\
		\hline
		4 prototype, 3 recurrence
		&82.1&88.1&82.7&\textbf{79.8}&92.8\\
		\hline
		\hline
		8 prototype, 1 recurrence
		&81.1&87.6&80.9&74.7&92.1\\
		\hline
		8 prototype, 2 recurrence
		&\textbf{82.5}&\textbf{88.7}&82.5&78.6&92.8\\
		\hline
		8 prototype, 3 recurrence
		&82.4&88.2&\textbf{83}&78.6&\textbf{92.9}\\
	\end{tabular}
	\caption{Comparison for different prototype numbers of TDAPNet on PASCAL3D+. We just list Level-3 and Mean here, the complete results are in the Supplementary Material.}
	\label{prototype_number}
\end{table}

\noindent\textbf{Comparison between TDAPNet and CompDictModel.} Dictionary-based Compositional Model \citep{kortylewski2019compositional} is a model that uses both VGG and a compositional model for classfication under partial occlusion. Details refer to Related Work. Results show that TDAPNet outperforms CompDictModel in most conditions, especially when occlusion level is low. However, there are some conditions that CompDictModel performs better. A possible reason is that CompDictModel requires a complex model of occlusion with some assumptions for occlusion. Differently, our model do not require assumptions for occlusion due to its possibly large variance and may be more generalizable to anomalous occlusion conditions. Another reason is that CompDictModel learns compositional models from the pool-4 layer, which benefits certain conditions. Details refer to Supplementary Material. In addition, our model is end-to-end, with fewer parameters and is more simple in computation.

\subsection{Comparison of prototype number}

\begin{table*} [htbp]
	\small
	\centering
	\tabcolsep=0.11cm
	\begin{tabular}{lV{2.5}c|c|c|cV{2.5}c|c|c|cV{2.5}c|c|c|c}
		\multicolumn{13}{c}{\textbf{Contamination Reduction on PASCAL3D+}} \\
		\hline
		Occ. Area 	& \multicolumn{4}{cV{2.5}}{\textbf{Level-1: 20-40\%}} & \multicolumn{4}{cV{2.5}}{\textbf{Level-2: 40-60\%}} & \multicolumn{4}{c}{\textbf{Level-3: 60-80\%}}\\
		\hline
		Occ. Type 	& w & n & t & o & w & n & t& o & w & n & t & o\\    		
		\hline  
		\hline
		TDAPNet with 1 recurrence
		&14.1\%&15.2\%&15.7\%&12.5\%&9.5\%&9.2\%&10.1\%&11.7\%&11.1\%&11.1\%&12.9\%&10.9\%\\
		\hline
		TDAPNet with 2 recurrence
		&16.7\%&17.9\%&18.6\%&13.9\%&10.3\%&10.0\%&10.9\%&13.1\%&13.0\%&13.1\%&15.3\%&12.1\%\\
		\hline
		TDAPNet with 3 recurrence
		&19.8\%&21.1\%&21.9\%&16.0\%&10.9\%&10.7\%&11.6\%&15.3\%&15.3\%&15.5\%&17.8\%&14.1\%\\
		\hline
		TDAPNet with 4 recurrence
		&19.9\%&21.4\%&22.2\%&15.8\%&10.6\%&10.4\%&11.4\%&15.3\%&15.6\%&15.8\%&18.2\%&14.2\%\\
	\end{tabular}
	\caption{Contamination reduction percentage by top-down regulation. Larger number reflects better results.}
	\label{contamination_reduction}
\end{table*}

In previous experiments, we set prototype number to be 4 to compare with CompDictModel under the same setting. As different prototypes in one class are designed to account for large variance in spatial patterns due to, for example, different viewpoints, we further compare different prototype numbers and visualize images that belong to the same prototype.

\noindent\textbf{Multiple prototypes improve the performance}. As shown in Table \ref{prototype_number}, 4 prototypes outperform 1 prototype, while 8 prototypes lead to about the same performance as 4 prototypes. It implies that modeling different spatial patterns enables prototypes to be more inclusive, and 4 prototypes are probably enough to account for the spatial variance in the PASCAL3D+ dataset.

\begin{figure} [htbp]
    \centering
    \subfigure[]{
    \includegraphics[scale=0.33]{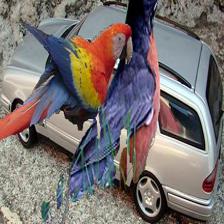}
    \label{image_occ}
    }
    \hspace{2mm}
    \subfigure[]{
    \includegraphics[scale=0.15]{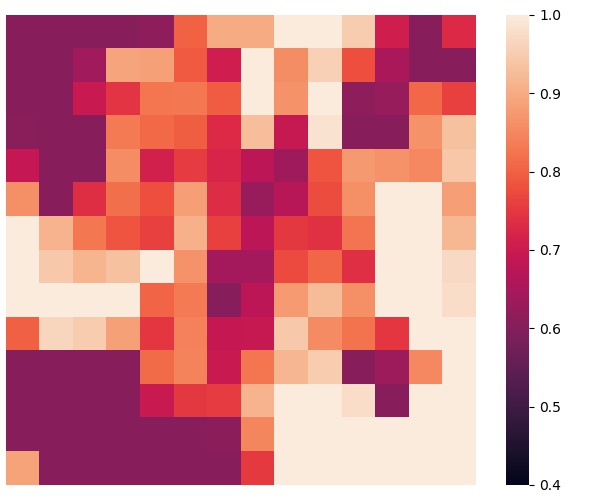}
    \label{attention_scale}
    }
    \subfigure[]{
    \includegraphics[scale=0.15]{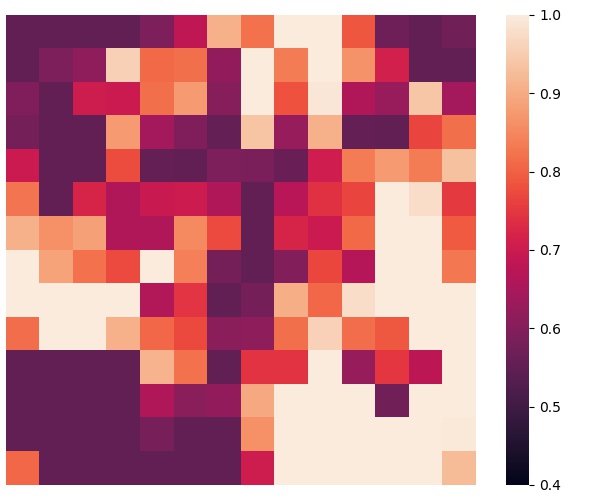}
    \label{attention}
    }
    \quad
    \subfigure[]{
    \includegraphics[scale=0.15]{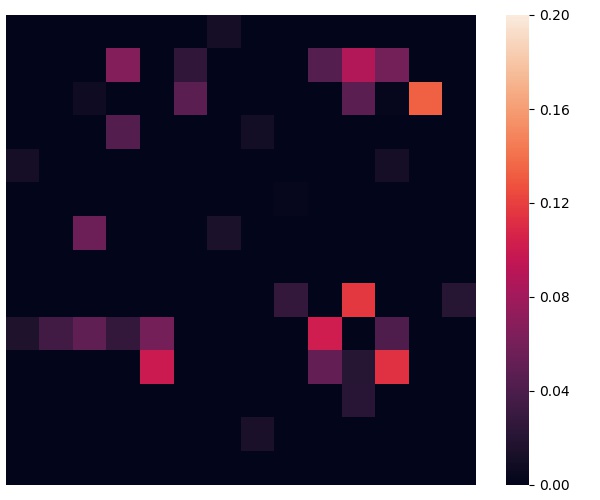}
    \label{attention_vc_attention}
    }
    \subfigure[]{
    \includegraphics[scale=0.15]{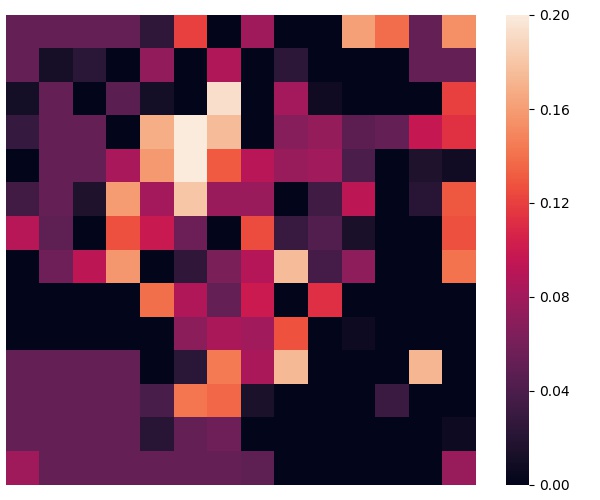}
    \label{attention_vc_filter}
    }
    \subfigure[]{
    \includegraphics[scale=0.15]{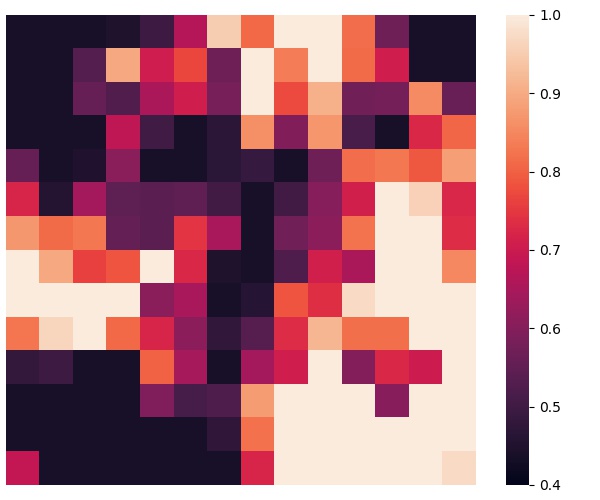}
    \label{attention_new}
    }
    \caption{Visualizaion of attention. Lighter areas represent more attention. (a) is the occluded image. (b) is the attention map only based on activation scale. (c) is the attention map based on feature dictionary and activation scale in the first feed-forward procedure. (d) is the attention map where feature dictionary enhance attention compared with (b). (e) is the attention map where feature dictionary reduce attention compared with (b). (f) is the final attention map after one top-down recurrence.}
    \label{attention visualization}
\end{figure}

\noindent\textbf{Multiple prototypes maintain spatial strutures}. As shown in the Supplementary Material, the four prototypes in our model mainly correspond to different viewpoints with certain spatial structure. When there's only one prototype for each class, what network learns is probably a metric that pushes all possible activation patterns of the feature of an object at a certain position to be close with each other, so that the prototype of its class is always the closest to it during matching. This may lose spatial structure of objects. Multiple prototypes is able to tackle such problem effectively.

\subsection{Analysis of the attention functioning}

The classification results demonstrate the importance of attention and partial matching. We illustrate how the two components in the attention contribute to focusing on informative parts through visualization of attention maps. As shown in the Figure \ref{attention visualization}, the activation scale (\ref{attention_scale}) reduce the attention to the background in \ref{image_occ}. Based on it the feature dictionary further reduce the attention on the occluding parrots (\ref{attention_vc_filter}) and enhance attention on several positions (\ref{attention_vc_attention}), resulting in attention map \ref{attention}. After a top-down recurrence, the attention further filters irrelevant information out and mainly focuses on informative parts (\ref{attention_new}).

\subsection{Analysis of the top-down recurrence effect}

\begin{figure} [htbp]
    \centering
    \subfigure[]{
    \includegraphics[scale=0.26]{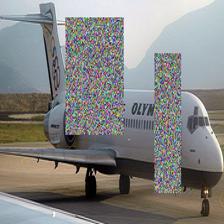}
    \label{0_occlusion}
    }
    \hspace{1mm}
    \subfigure[]{
    \includegraphics[scale=0.12]{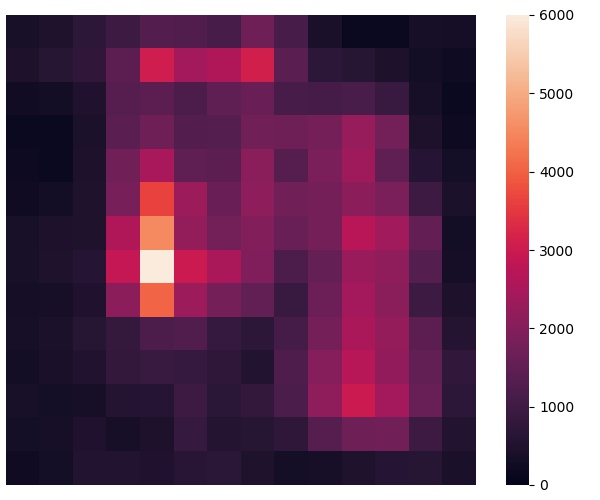}
    \label{0_diff11}
    }
    \subfigure[]{
    \includegraphics[scale=0.12]{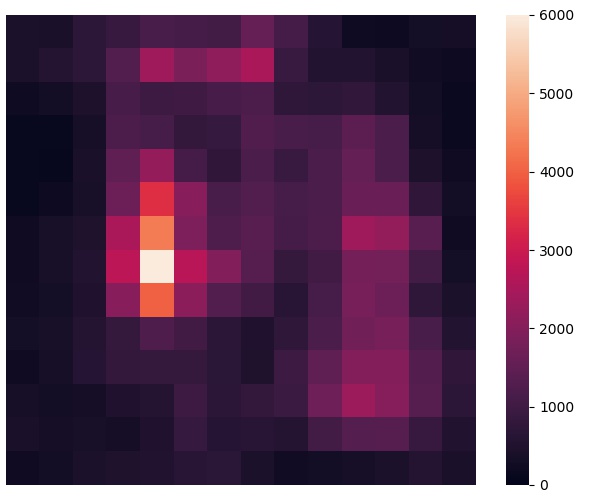}
    \label{0_diff22}
    }
    \subfigure[]{
    \includegraphics[scale=0.12]{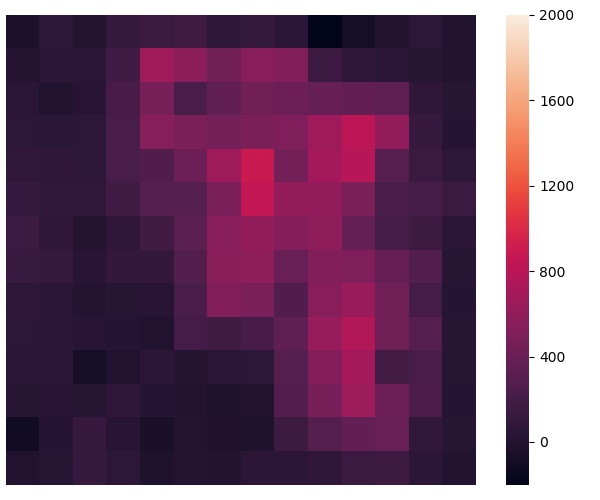}
    \label{0_diff_diff}
    }
    \quad
    \subfigure[]{
    \includegraphics[scale=0.26]{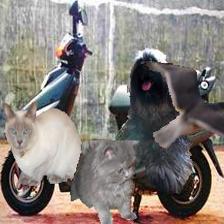}
    \label{2_occlusion}
    }
    \hspace{1mm}
    \subfigure[]{
    \includegraphics[scale=0.12]{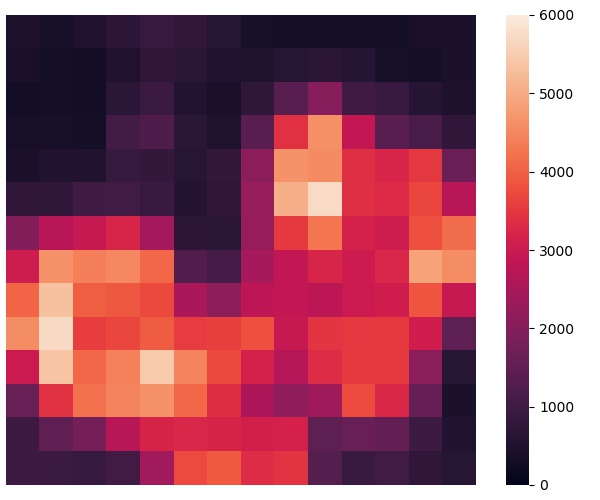}
    \label{2_diff11}
    }
    \subfigure[]{
    \includegraphics[scale=0.12]{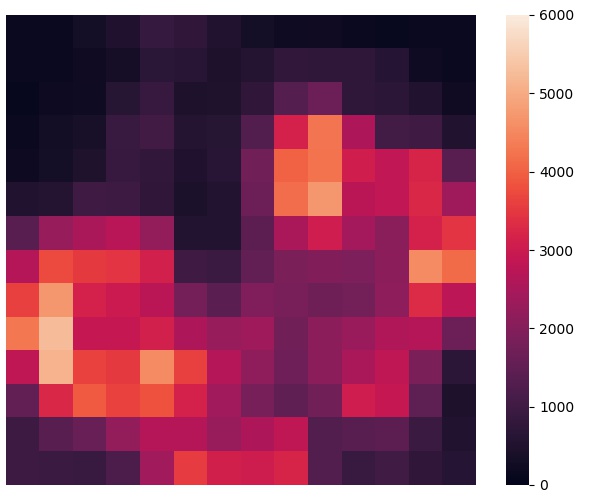}
    \label{2_diff22}
    }
    \subfigure[]{
    \includegraphics[scale=0.12]{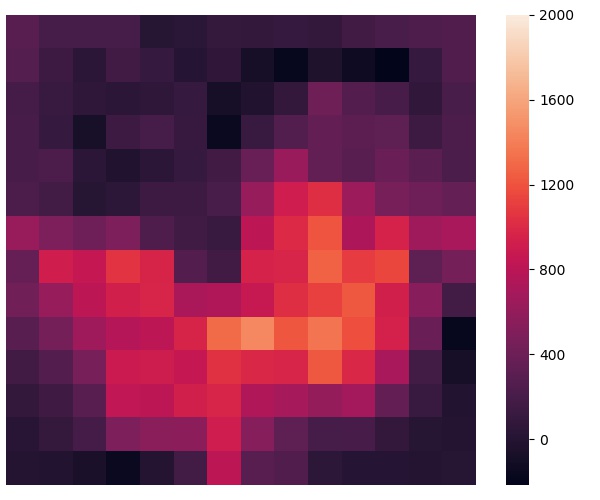}
    \label{2_diff_diff}
    }
    \caption{Visualizaion of difference reduction after one top-down recurrence at pool-4 layer. (a)\&(e) are the occluded images. (b)\&(f) are the activation difference between clean and occluded images before top-down recurrence, while (c)\&(g) are the activation difference after top-down recurrence. Lighter areas represent more difference. (d)\&(h) are the difference reduction. Lighter areas represent more difference reduction.}
    \label{difference_reduction}
\end{figure}

We further validate the function of recurrent top-down attention regulation. It is designed to reduce contamination of occlusion to its surroundings. Therefore, we compare the differences between the pool-4 feature of the clean images and the occluded images before and after top-down recurrence. Specifically, let $f_c^0$, $f_c^r$, $f_o^0$, $f_o^r$ denote the pool-4 feature of the clean image before and after recurrence and the occluded image before and after recurrence respectively, and let $m_o$ denote the mask of occlusion area obtained by average down-sampling of the occlusion ground truth. We compute $R_c=1-\frac{sum(|f_o^r \odot m_o - f_c^r \odot m_o|)}{sum(|f_o^1 \odot m_o - f_c^1 \odot m_o|)}$ as the contamination reduction percentage. The results are illustrated in Table \ref{contamination_reduction}. It clearly shows that top-down recurrence is capable of reducing contamination in the bottom layer based on attention from the top layer, and nearly the more the recurrence, the more the reduction. Further, the visualization of difference reduction is shown in Figure \ref{difference_reduction}. It shows that the top-down recurrence is capable of reducing occlusion-caused difference in features surrounding the occlusions.



\section{Conclusion}

In this work, we integrate prototype learning, partial matching based on informative-part attention, and top-down attention regulation into deep neural networks for robust object classification under occlusion. The estimated attention could effectively extend prototype matching to occlusion conditions, and the top-down regulation based on the estimated attention could deal with the contamination of occlusion during feature extraction. Our model significantly improves current deep networks, and is possible for extension to various network structures.

\bibliography{reference}
\bibliographystyle{aaai}


\section{Supplementary Material}
\vspace{1.5cm}

\subsection{Comparison of different prototype number}

\begin{table*} [htp]
	\begin{center}
	\tabcolsep=0.11cm
	\begin{tabular}{lV{2.5}cV{2.5}c|c|c|cV{2.5}c|c|c|cV{2.5}c|c|c|cV{2.5}c}
		\multicolumn{15}{c}{\textbf{PASCAL3D+ Classification under Occlusion}} \\
		\hline
		Occ. Area 					         & \textbf{0\%} & \multicolumn{4}{cV{2.5}}{\textbf{Level-1: 20-40\%}} & \multicolumn{4}{cV{2.5}}{\textbf{Level-2: 40-60\%}} & \multicolumn{4}{cV{2.5}}{\textbf{Level-3: 60-80\%}} & Mean \\
		\hline
		Occ. Type 	& - & w & n & t & o & w & n & t& o & w & n & t & o &-\\    		
		\hline		
		\hline
		1 prototype, 1 recurrence
		&99.2&97.9&98.5&97.9&96.4&95.1&96.5&95.2&88.6&79.1&84.8&77.9&75.1&90.9\\
		\hline					
		1 prototype, 2 recurrence
		&99.2&97.9&98.3&97.9&96.3&95.1&96.6&95.0&89.1 &78.9&85.2&78.2&75.5&91.0\\
		\hline
		1 prototype, 3 recurrence
		&99.0&98.0&98.3&97.8&96.5&94.7&96.3&95.3&89.5&79.7&85.2&79.0&76.9&91.2\\
		\hline		
		\hline
		4 prototype, 1 recurrence
		&99.3&98.4&98.9&\textbf{98.7}&97.2&96.1&97.4&96.4&90.2&81.1&87.6&81.2&76.8&92.3\\
		\hline					
		4 prototype, 2 recurrence
		&99.2&98.5&98.8&98.5&97.3&96.2&97.4&96.6&90.2&81.5&87.7&81.9&77.1&92.4\\
		\hline
		4 prototype, 3 recurrence
		&99.3&98.4&98.9&98.5&97.4&96.1&\textbf{97.5}&96.6&\textbf{91.6}&82.1&88.1&82.7&\textbf{79.8}&92.8\\
		\hline		
		\hline
		8 prototype, 1 recurrence
		&99.3&\textbf{98.7}&98.9&\textbf{98.7}&97.5&\textbf{96.4}&\textbf{97.5}&96.7&89.6&81.1&87.6&80.9&74.7&92.1\\
		\hline					
		8 prototype, 2 recurrence
		&\textbf{99.4}&\textbf{98.7}&99.0&98.6&\textbf{97.7}&96.1&\textbf{97.5}&\textbf{96.8}&90.8&\textbf{82.5}&\textbf{88.7}&82.5&78.6&92.8\\
		\hline
		8 prototype, 3 recurrence
		&99.3&98.6&\textbf{99.1}&98.6&97.6&96.2&\textbf{97.5}&96.7&91.4&82.4&88.2&\textbf{83.0}&78.6&\textbf{92.9}
	\end{tabular}
	\end{center}
	\caption{Comparison of different prototype numbers for TDAPNet on PASCAL3D+. It shows that 4 prototypes outperform 1 prototype in every condition, while 8 prototypes lead to about the same performance as 4 prototypes, with a slight improvement.}
	\label{supplementary prototype number}
\end{table*}

Shown in Table \ref{supplementary prototype number}.

\subsection{Comparison of prototype learning on different layers}

\begin{table*} [htp]
	\begin{center}
	\tabcolsep=0.11cm
	\begin{tabular}{lV{2.5}cV{2.5}c|c|c|cV{2.5}c|c|c|cV{2.5}c|c|c|cV{2.5}c}
		\multicolumn{15}{c}{\textbf{PASCAL3D+ Classification under Occlusion}} \\
		\hline
		Occ. Area 					         & \textbf{0\%} & \multicolumn{4}{cV{2.5}}{\textbf{Level-1: 20-40\%}} & \multicolumn{4}{cV{2.5}}{\textbf{Level-2: 40-60\%}} & \multicolumn{4}{cV{2.5}}{\textbf{Level-3: 60-80\%}} & Mean \\
		\hline
		Occ. Type 	& - & w & n & t & o & w & n & t& o & w & n & t & o &-\\  
		\hline  
		\hline
		VGG 		& \textbf{99.4} &97.5&97.5&97.3&92.1&91.7&90.6&90.2&73.0&65.0&60.7&56.4&52.2&81.8\\	
		\hline
		CompDictModel	&98.3&96.8&95.9&96.2&94.4&91.2&91.8&91.3&91.4&71.6&80.7&77.3&\textbf{87.2}&89.5\\
		\hline		
		\hline
		pool-5, without recurrence
		&99.3&\textbf{98.4}&\textbf{98.9}&98.5&\textbf{97.3}&\textbf{96.4}&97.1&96.2&89.2&\textbf{84.0}&87.4&79.7&74.5&92.1\\
		\hline
		pool-5, with 1 recurrence
		&99.3&\textbf{98.4}&\textbf{98.9}&\textbf{98.7}&97.2&96.1&\textbf{97.4}&\textbf{96.4}&90.2&81.1&\textbf{87.6}&\textbf{81.2}&76.8&\textbf{92.3}\\
		\hline	
		\hline
		pool-4, without recurrence
		&98.1&97.6&98.0&97.8&96.4&94.8&96.3&95.2&\textbf{92.0}&76.5&85.2&79.5&84.2&91.7\\
		\hline
		pool-4, with 1 recurrence
		&98.0&97.3&97.9&97.8&96.0&94.6&96.1&95.3&91.7&77.5&85.9&80.5&84.4&91.8\\
	\end{tabular}
	\end{center}
	\caption{Comparison of prototype learning on pool-4 layer and pool-5 layer for TDAPNet on PASCAL3D+. It shows that prototype learning on pool-5 layer outperforms pool-4 layer in zero or low occlusion conditions and when occlusion is white boxes, noise boxes, or textures, while prototype learning on pool-4 layer performs well when occlusion is object under high occlusion level.}
	\label{pool}
\end{table*}

As discussed in the experiment part, CompDictModel outperforms TDAPNet in some conditions. A difference between CompDictModel and TDAPNet is that TDAPNet learn prototypes from the pool-5 layer while CompDictModel learn compositional model from the pool-4 layer. We further conduct experiments to learn prototypes from the pool-4 layer, as shown in Table \ref{pool}. The results show that performance drops under most conditions while increases under object occlusion at level-3 if we learn prototypes from the pool-4 layer. It performs similar with CompDictModel. A possible reason is that the information in the pool-4 layer is more local and part-based, so it does not perform as well as information from the pool-5 layer under zero or low occlusion conditions. But under severe occlusion with irregular shape, local information may play a more important role for object recognition, leading to its high performance under the certain condition. A mechanism to combine the information from the pool-4 layer and the pool-5 layer might utilize both of their advantages.

\subsection{Visualization of different prototypes}
\begin{figure*} [!ht]
    \centering
    \subfigure[]{
    \includegraphics[scale=0.125]{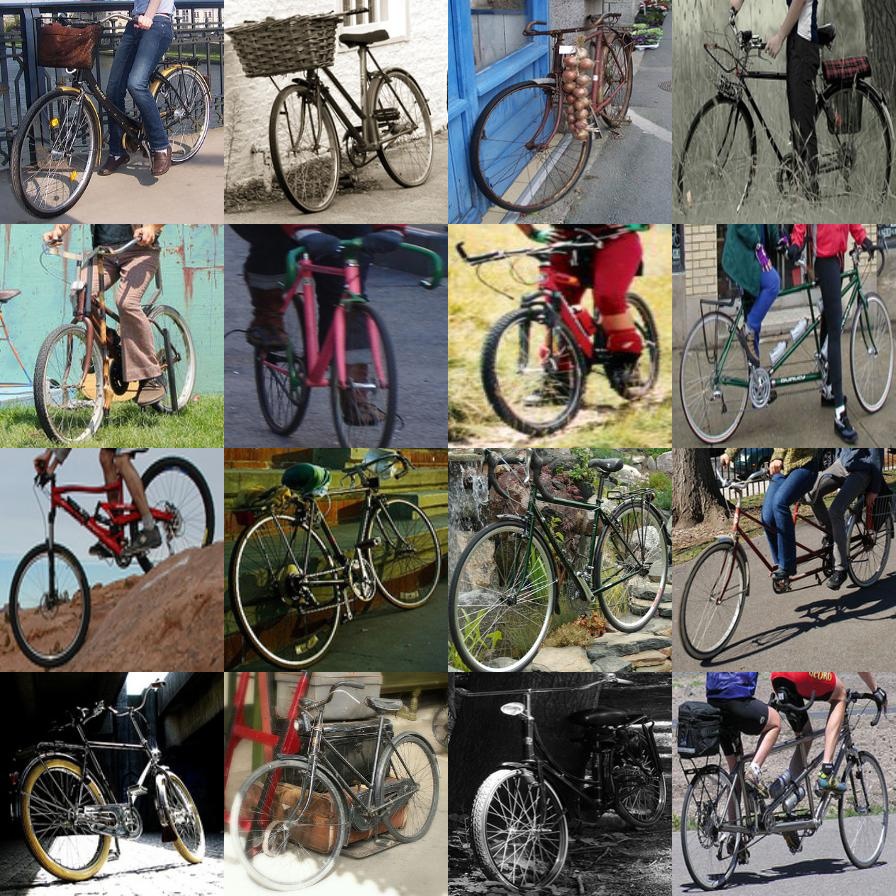}
    \label{1_0}
    }
    \subfigure[]{
    \includegraphics[scale=0.125]{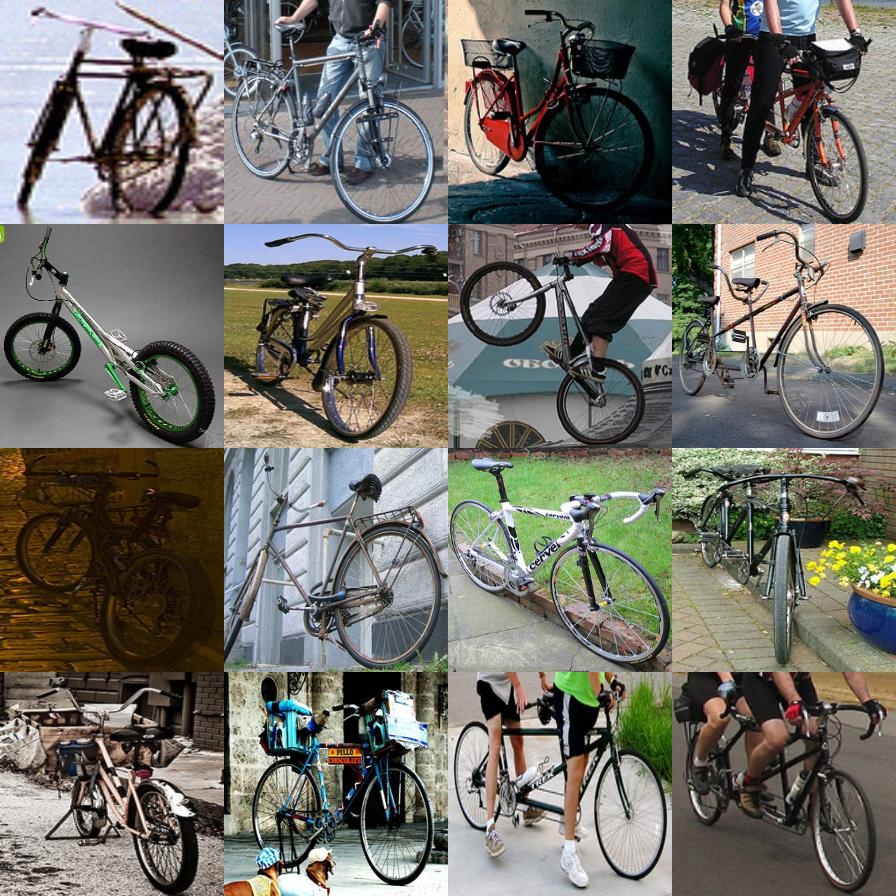}
    \label{1_1}
    }
    \subfigure[]{
    \includegraphics[scale=0.125]{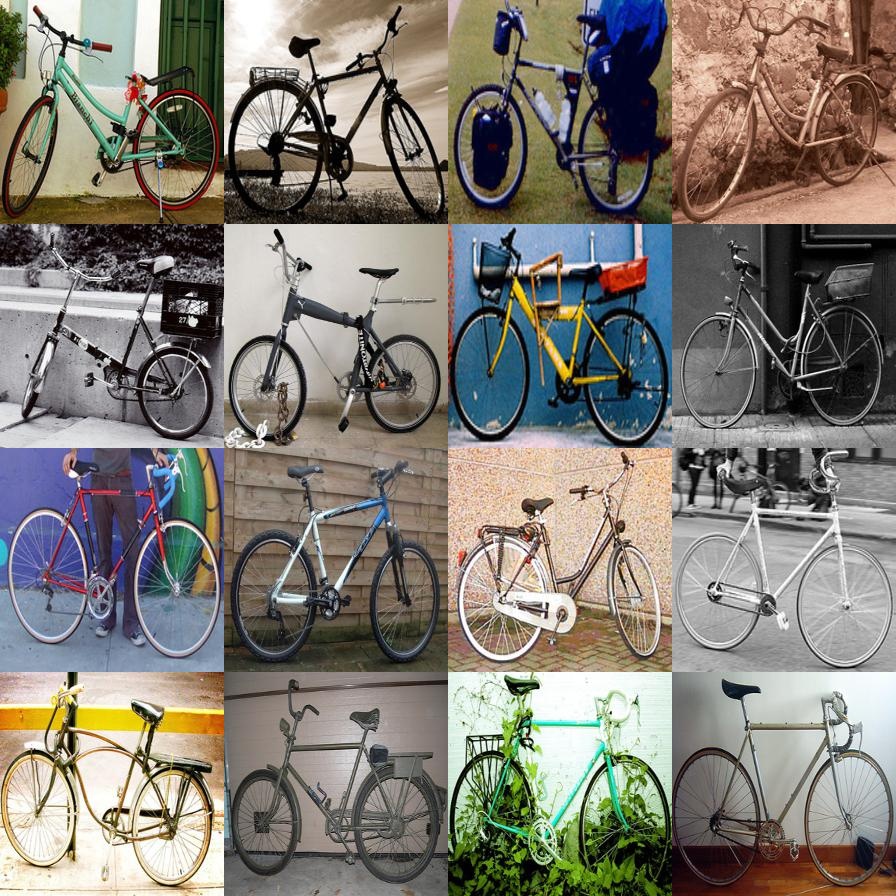}
    \label{1_2}
    }
    \subfigure[]{
    \includegraphics[scale=0.125]{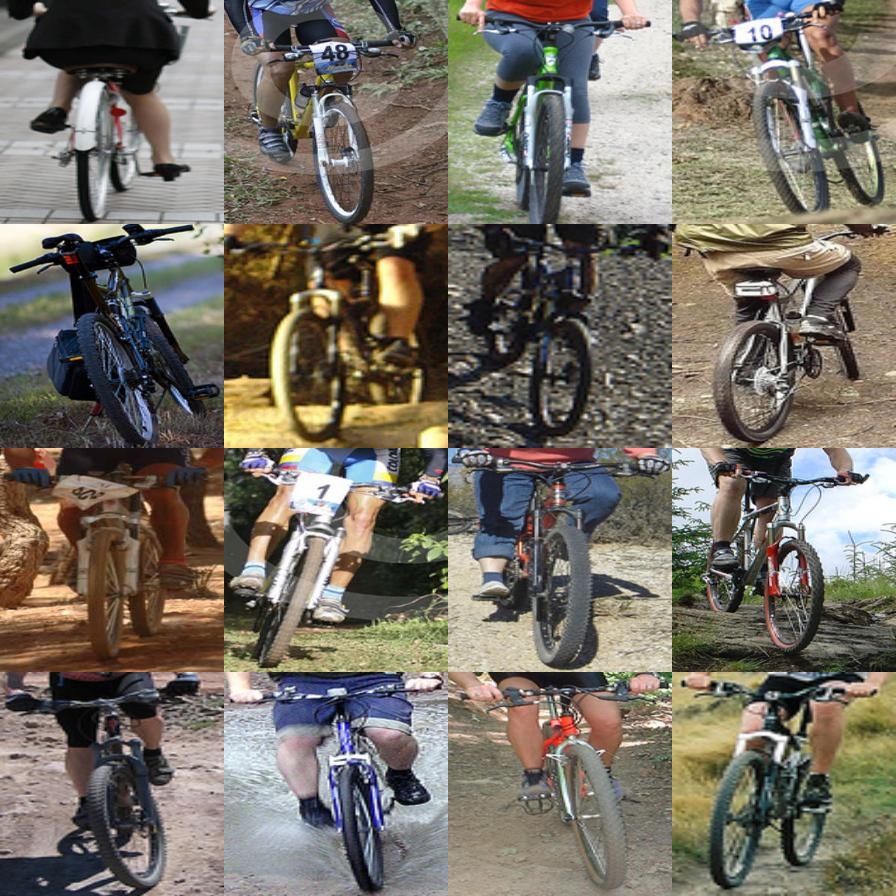}
    \label{1_3}
    }
    \quad
    \subfigure[]{
    \includegraphics[scale=0.125]{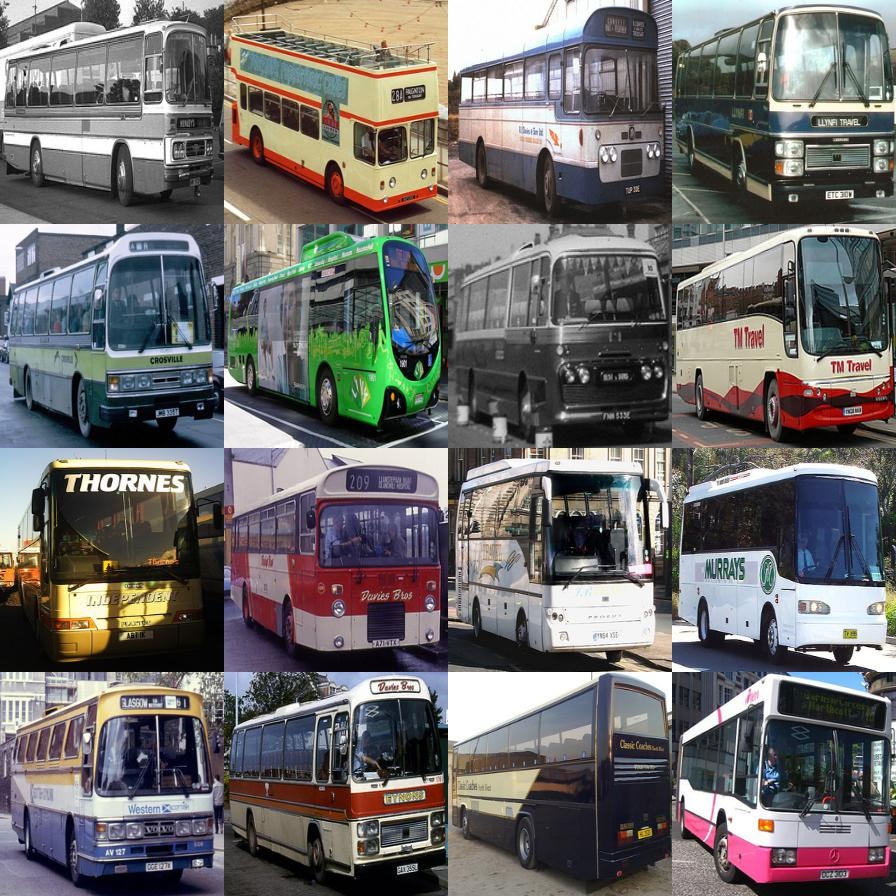}
    \label{4_0}
    }
    \subfigure[]{
    \includegraphics[scale=0.125]{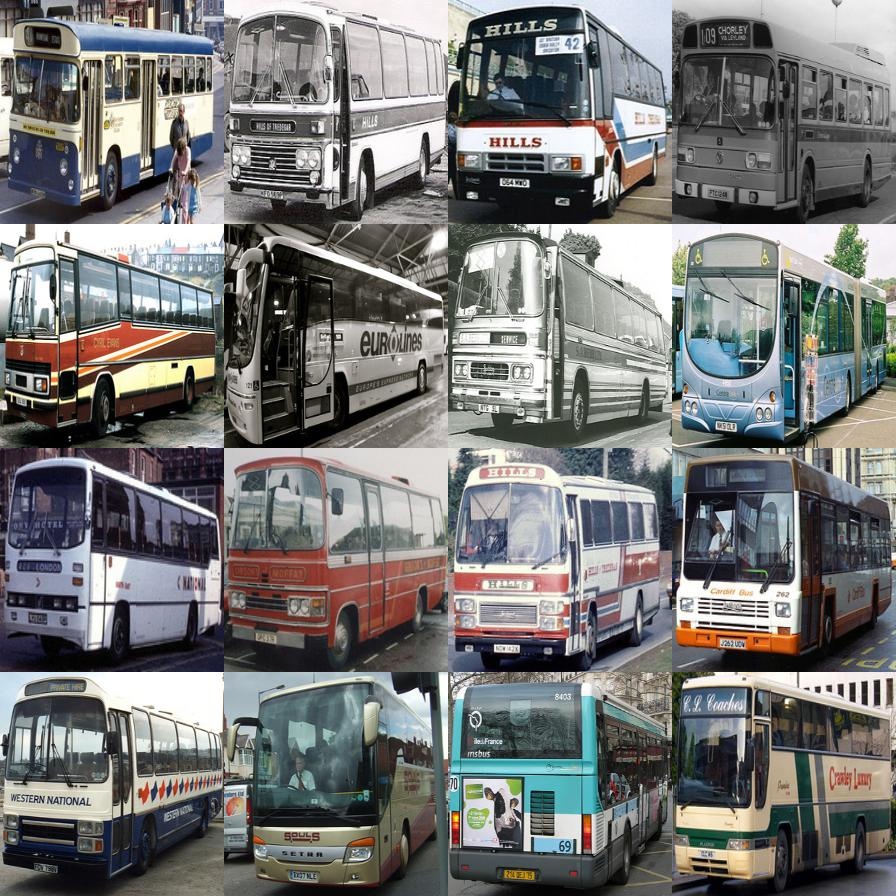}
    \label{4_1}
    }
    \subfigure[]{
    \includegraphics[scale=0.125]{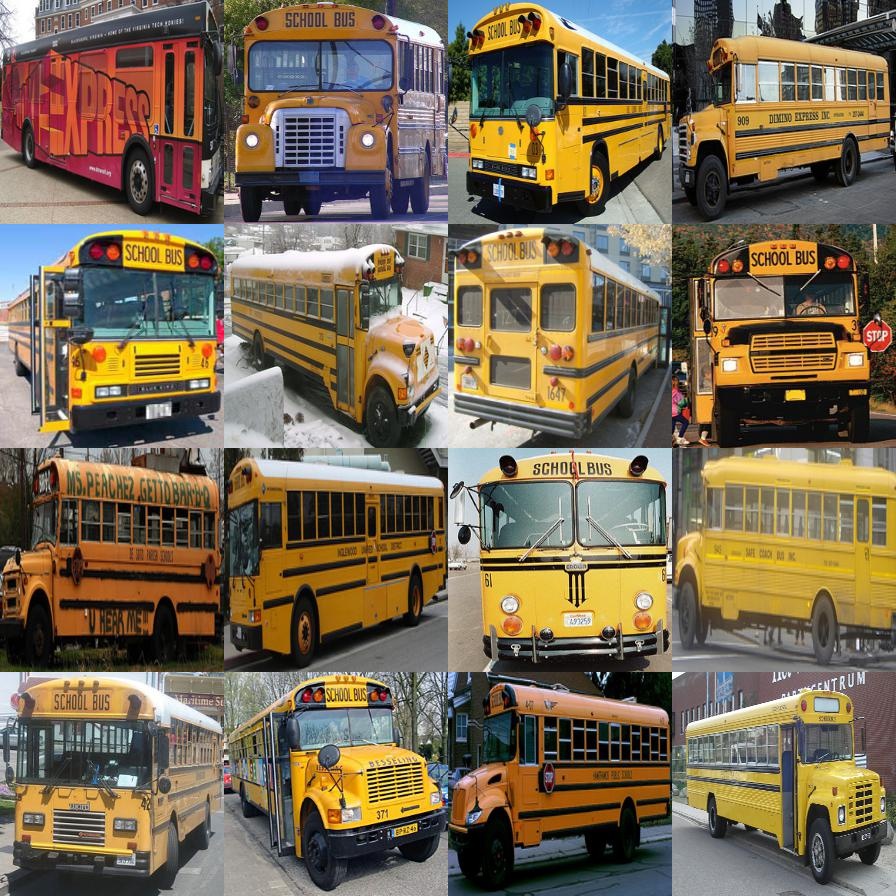}
    \label{4_2}
    }
    \subfigure[]{
    \includegraphics[scale=0.125]{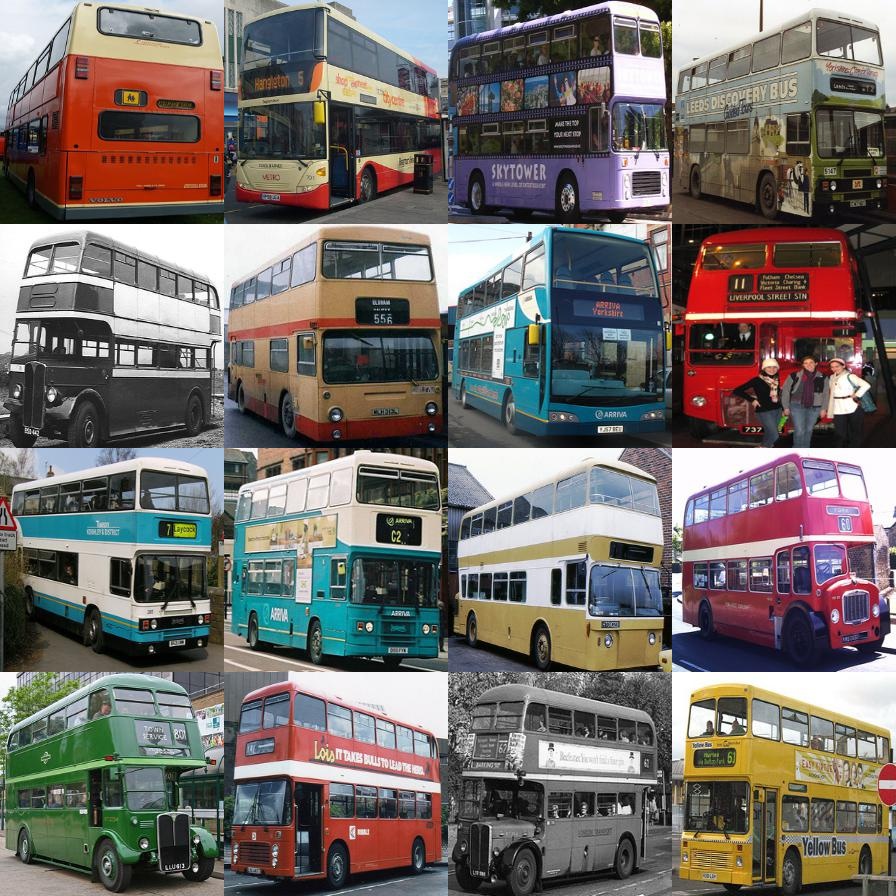}
    \label{4_3}
    }
    \quad
    \subfigure[]{
    \includegraphics[scale=0.125]{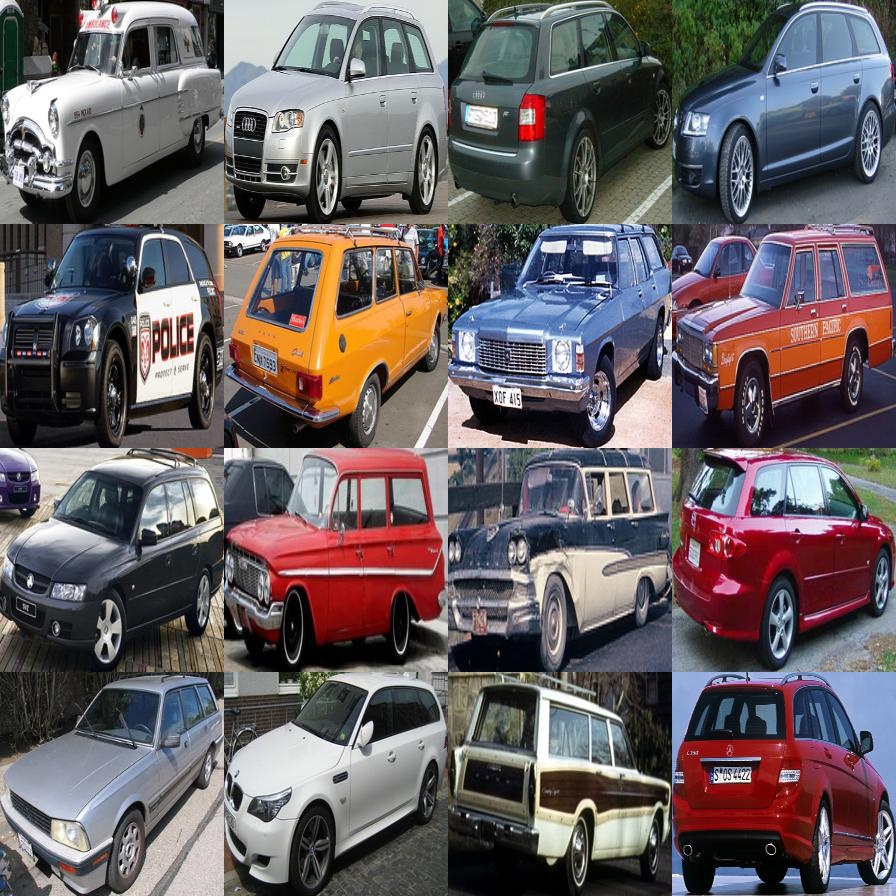}
    \label{5_0}
    }
    \subfigure[]{
    \includegraphics[scale=0.125]{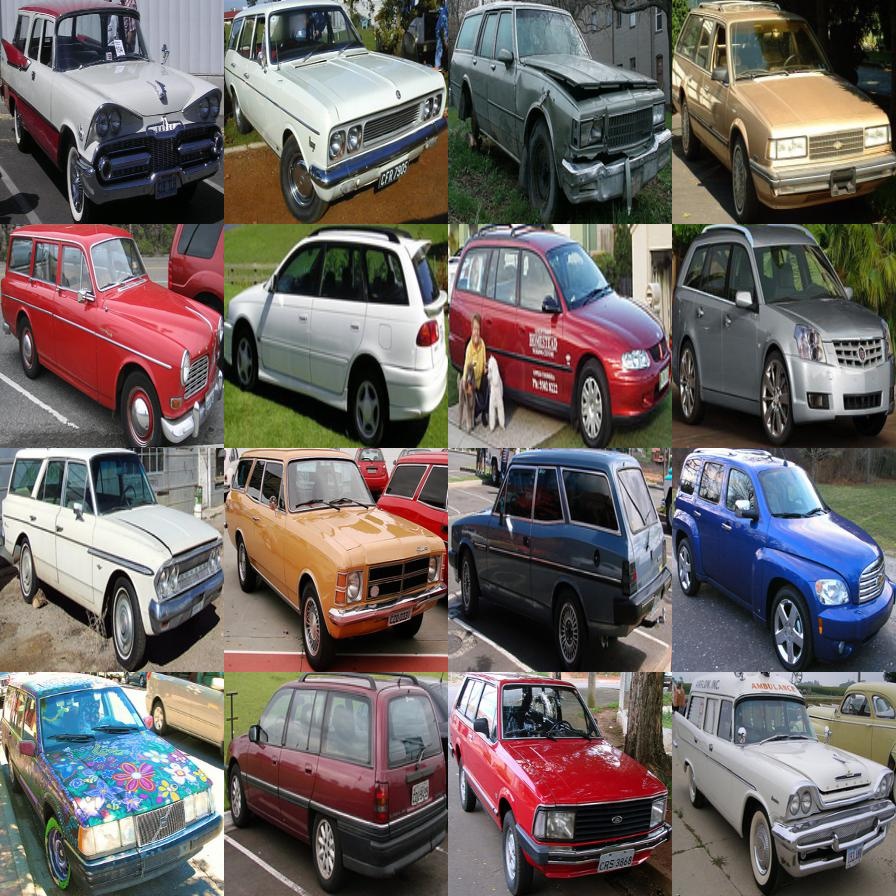}
    \label{5_1}
    }
    \subfigure[]{
    \includegraphics[scale=0.125]{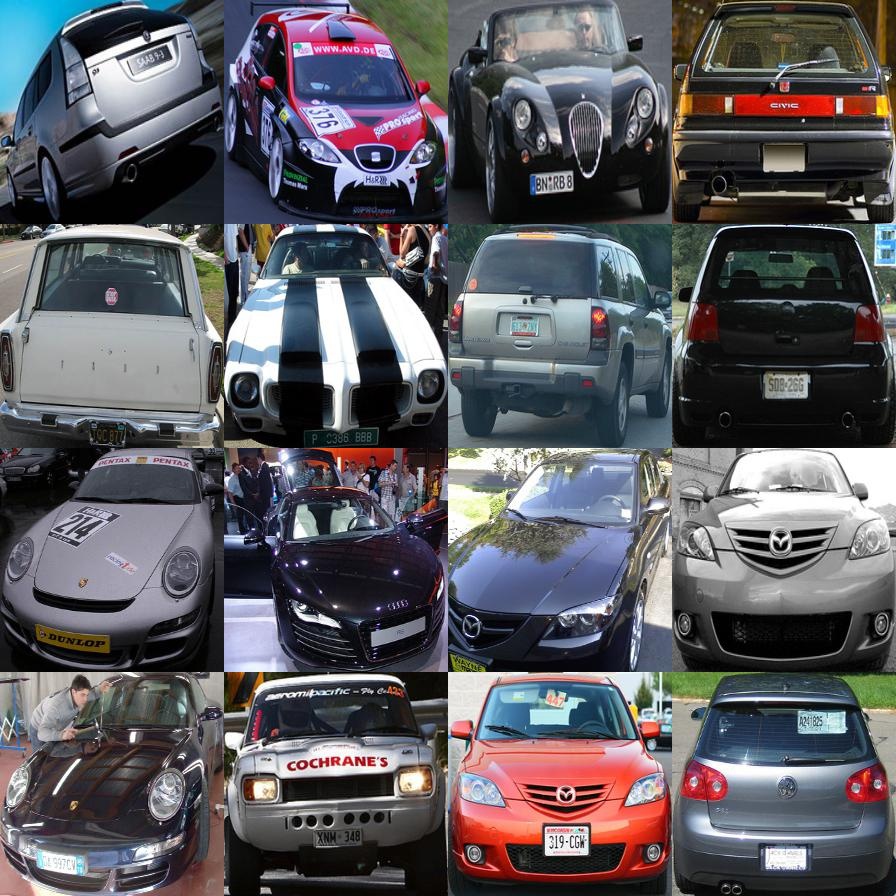}
    \label{5_2}
    }
    \subfigure[]{
    \includegraphics[scale=0.125]{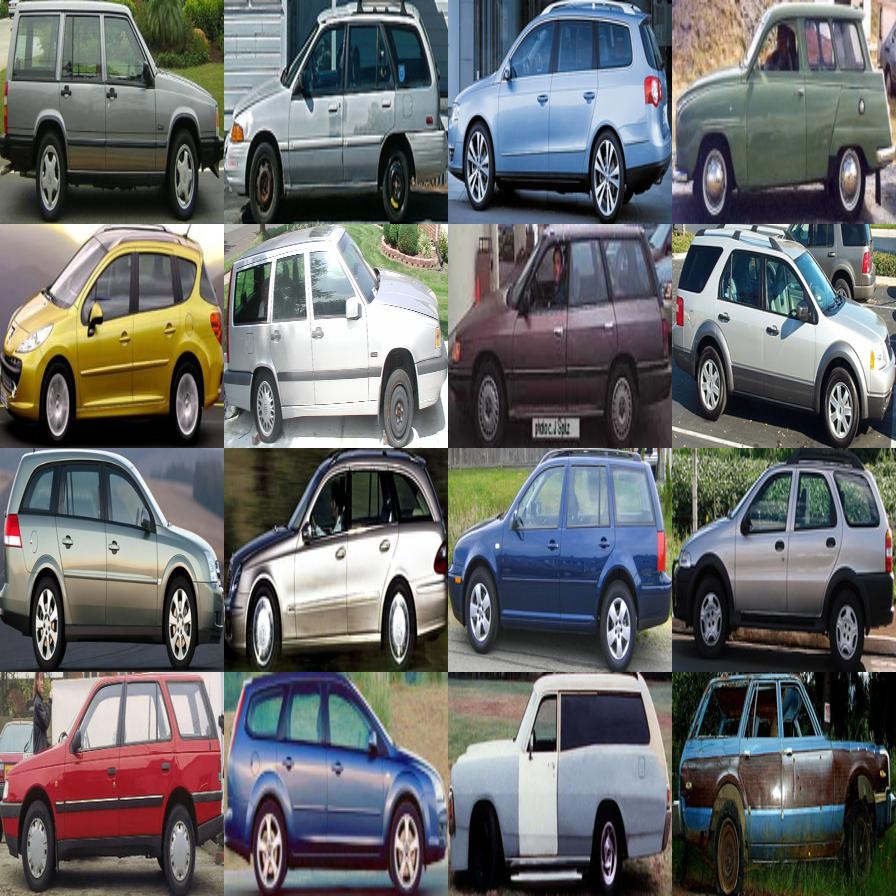}
    \label{5_3}
    }
    \quad
    \subfigure[]{
    \includegraphics[scale=0.125]{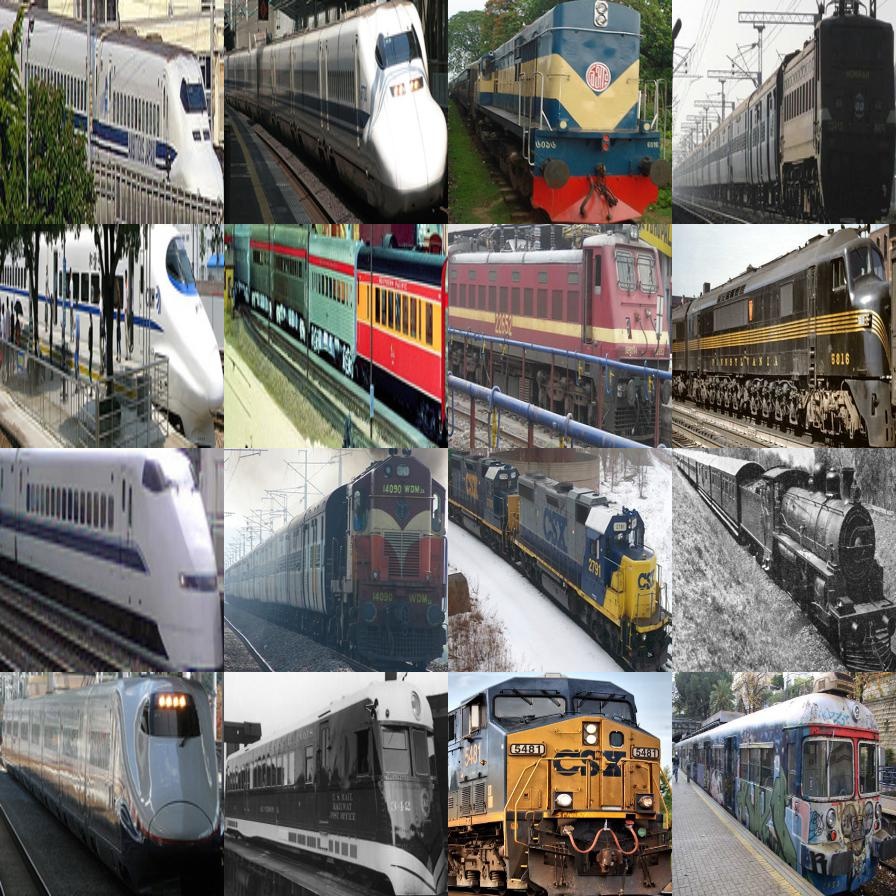}
    \label{10_0}
    }
    \subfigure[]{
    \includegraphics[scale=0.125]{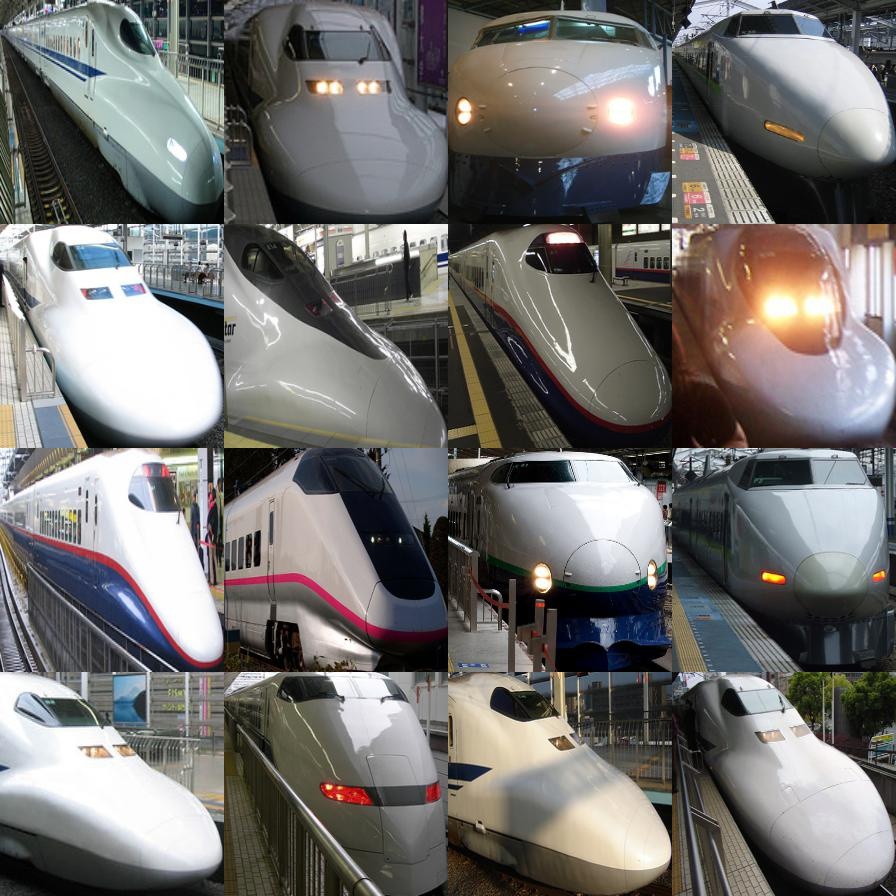}
    \label{10_1}
    }
    \subfigure[]{
    \includegraphics[scale=0.125]{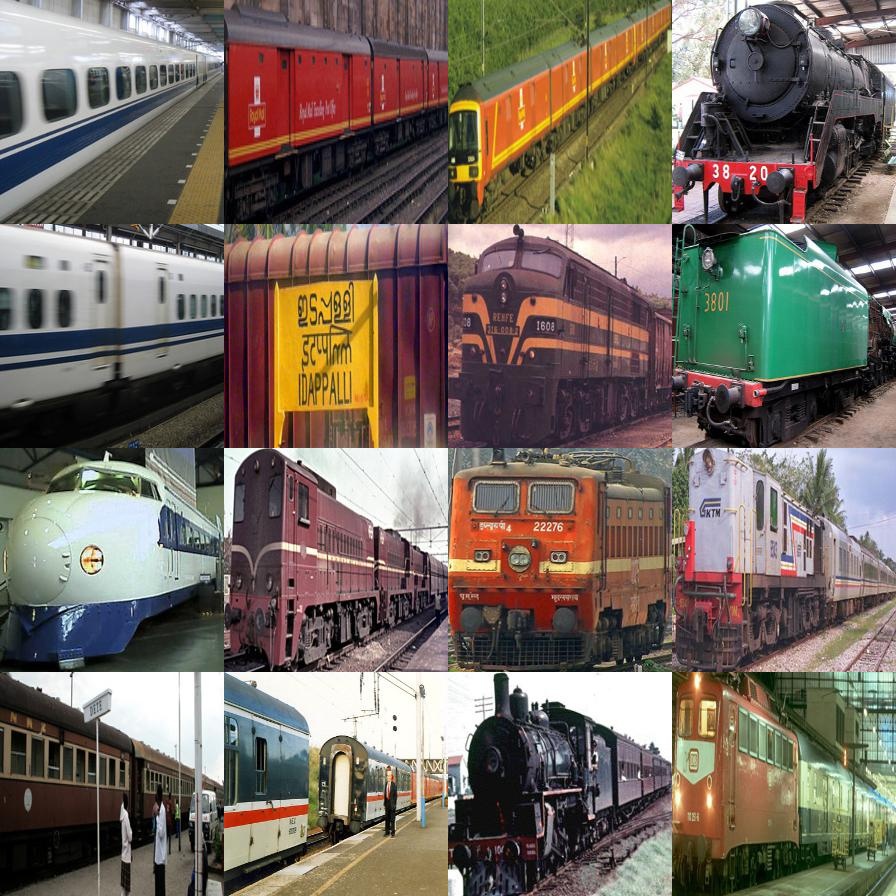}
    \label{10_2}
    }
    \subfigure[]{
    \includegraphics[scale=0.125]{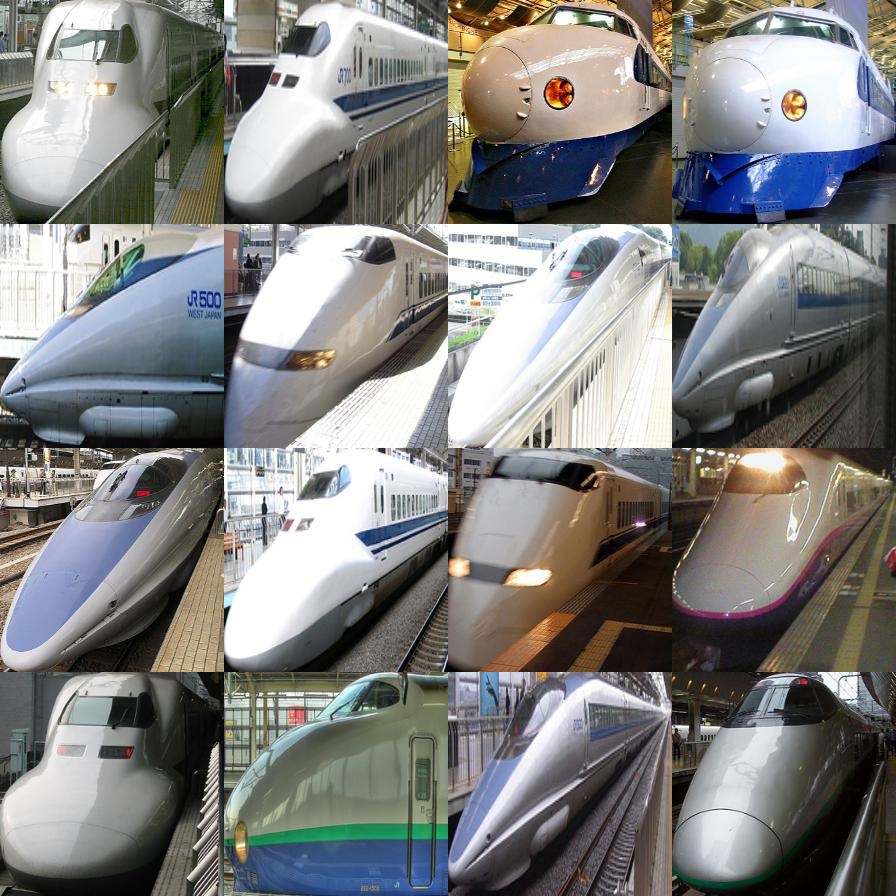}
    \label{10_3}
    }
    \quad
    \caption{Visualization of different prototypes. It shows that different prototypes could mainly account for different spatial distribution caused by viewpoints. And if there are more prototypes than viewpoints, some prototypes could focus on some specific features, like school bus in (g) and double-decker bus in (h), or different appearance shown in (m), (n), (o), and (p)}
    \label{prototype visualization}
\end{figure*}

Shown in Figure \ref{prototype visualization}.

\end{document}